\documentclass{article}

\usepackage{natbib}

\usepackage{hyperref}

\usepackage[accepted]{icml2014}

\usepackage[utf8]{inputenc}

\usepackage{amsmath,amssymb, amsfonts}
\usepackage{bm,color,epsf,epsfig,graphicx,multirow,paralist,times,subfigure}
\usepackage{algorithm,algorithmic,bbm}

\newcommand{\EE}{\mathbb{E}}
\newcommand{\NN}{\mathbb{N}}
\newcommand{\PP}{\mathbb{P}}
\newcommand{\RR}{\mathbb{R}}
\newcommand{\indic}{{\bf 1}}

\newcommand{\bp}{\noindent{\bf Proof.}\ }
\newcommand{\ep}{\hfill $\Box$}

\newcommand{\al}[1]{ \begin{align} #1  \end{align}}
\newcommand{\eq}[1]{ \begin{equation} #1  \end{equation}}
\newcommand{\als}[1]{ \begin{align*} #1  \end{align*}}
\newcommand{\eqs}[1]{ \begin{equation*} #1  \end{equation*}}

\newcommand{\sk}{\nonumber\\} 
\newcommand{\Lp}{\left(}
\newcommand{\Rp}{\right)}
\newcommand{\Lb}{\left[}
\newcommand{\Rb}{\right]}

\newcommand{\el}{\end{flushleft}}
\newcommand{\bl}{\begin{flushleft}}
\newcommand{\floor}[1]{  \lfloor #1 \rfloor} 
\newcommand{\ceil}[1]{  \lceil #1 \rceil} 

\newcommand{\ouralgo}{OSUB } 
\newcommand{\ouralgosw}{SW-OSUB } 

\newcommand{\separator}{
  \begin{center}
    \rule{\columnwidth}{0.3mm}
  \end{center}
}
\newenvironment{separation}{ \vspace{-0.3cm}  \separator  \vspace{-0.2cm}}
{  \vspace{-0.4cm}  \separator  \vspace{-0.1cm}}

\newtheorem{proposition}{Proposition}
\newtheorem{theorem}{Theorem}[section]
\newtheorem{lemma}[theorem]{Lemma}
\newtheorem{corollary}[theorem]{Corollary}

\newtheorem{assumption}{Assumption}

\DeclareGraphicsExtensions{.pdf,.eps}

\graphicspath{{.//}}

\icmltitlerunning{Unimodal bandits}

\begin{document} 

\twocolumn[
\icmltitle{Unimodal Bandits:\texorpdfstring{\\}{ }Regret Lower Bounds and Optimal Algorithms}

\icmlauthor{Richard Combes}{rcombes@kth.se}
\icmladdress{KTH, Royal Institute of technology,
            Stockholm, Sweden}
\icmlauthor{Alexandre Proutiere}{alepro@kth.se}
\icmladdress{KTH, Royal Institute of technology,
            Stockholm, Sweden}
\icmlkeywords{Online Optimization, Multi-armed bandits, Unimodal bandits, Machine learning}

\vskip 0.3in
]

\begin{abstract}
We consider stochastic multi-armed bandits where the expected reward is a unimodal function over partially ordered arms. This important class of problems has been recently investigated in \cite{cope09, yu2011}. The set of arms is either discrete, in which case arms correspond to the vertices of a finite graph whose structure represents similarity in rewards, or continuous, in which case arms belong to a bounded interval. For discrete unimodal bandits, we derive asymptotic lower bounds for the regret achieved under any algorithm, and propose OSUB, an algorithm whose regret matches this lower bound. Our algorithm optimally exploits the unimodal structure of the problem, and surprisingly, its asymptotic regret does not depend on the number of arms. We also provide a regret upper bound for \ouralgo in non-stationary environments where the expected rewards smoothly evolve over time. The analytical results are supported by numerical experiments showing that \ouralgo performs significantly better than the state-of-the-art algorithms. For continuous sets of arms, we provide a brief discussion. We show that combining an appropriate discretization of the set of arms with the UCB algorithm yields an order-optimal regret, and in practice, outperforms recently proposed algorithms designed to exploit the unimodal structure.
\end{abstract}

\section{Introduction}\label{sec:introduction}

Stochastic Multi-Armed Bandits (MAB) \cite{robbins1952,gittins89} constitute the most fundamental sequential decision problems with an exploration vs. exploitation trade-off. In such problems, the decision maker selects an arm in each round, and observes a realization of the corresponding unknown reward distribution. Each decision is based on past decisions and observed rewards. The objective is to maximize the expected cumulative reward over some time horizon by balancing exploitation (arms with higher observed rewards should be selected often) and exploration (all arms should be explored  to learn their average rewards). Equivalently, the performance of a decision rule or algorithm can be measured through its expected {\it regret}, defined as the gap between the expected reward achieved by the algorithm and that achieved by an oracle algorithm always selecting the best arm. MAB problems have found many fields of application, including sequential clinical trials, communication systems, economics, see e.g. \cite{CesaBianchi06,bubeck12}. 

In their seminal paper \cite{lai1985}, Lai and Robbins solve MAB problems where the successive rewards of a given arm are i.i.d., and where the expected rewards of the various arms are not related. They derive an asymptotic (when the time horizon grows large) lower bound of the regret satisfied by any algorithm, and present an algorithm whose regret matches this lower bound. This initial algorithm was quite involved, and many researchers have tried to devise simpler and yet efficient algorithms. The most popular of these algorithms are UCB \cite{auer2002} and its extensions, e.g. KL-UCB \cite{garivier2011,cappe2012} (note that KL-UCB algorithm was initially proposed in \cite{lai1987}, see (2.6)). When the expected rewards of the various arms are not related \cite{lai1985}, the regret of the best algorithm is essentially of the order $O(K\log(T))$ where $K$ denotes the number of arms, and $T$ is the time horizon. When $K$ is very large or even infinite, MAB problems become more challenging. Fortunately, in such scenarios, the expected rewards often exhibit some structural properties that the decision maker can exploit to design efficient algorithms. Various structures have been investigated in the literature, e.g., Lipschitz \cite{agrawal95, kleinberg2008, bubeck08}, linear \cite{dani08}, convex \cite{kalai05}. 

We consider bandit problems where the expected reward is a unimodal function over partially ordered arms as in \cite{yu2011}. The set of arms is either discrete, in which case arms correspond to the vertices of a finite graph whose structure represents similarity in rewards, or continuous, in which case arms belong to a bounded interval. This unimodal structure occurs naturally in many practical decision problems, such as sequential pricing \cite{yu2011} and bidding in online sponsored search auctions \cite{edelman05}.   

{\bf Our contributions.} We mainly investigate unimodal bandits with finite sets of arms, and are primarily interested in cases where the time horizon $T$ is much larger than the number of arms $K$. 

(a) For these problems, we derive an asymptotic regret lower bound satisfied by any algorithm. This lower bound does not depend on the structure of the graph, nor on its size: it actually corresponds to the regret lower bound in a classical bandit problem \cite{lai1985}, where the set of arms is just a neighborhood of the best arm in the graph. 

(b) We propose OSUB (Optimal Sampling for Unimodal Bandits), a simple algorithm whose regret matches our lower bound, i.e., it optimally exploits the unimodal structure. The asymptotic regret of \ouralgo does not depend on the number of arms. This contrasts with LSE (Line Search Elimination), the algorithm proposed in \cite{yu2011} whose regret scales as $O(\gamma D\log(T))$ where $\gamma$ is the maximum degree of vertices in the graph and $D$ is its diameter. We present a finite-time analysis of OSUB, and derive a regret upper bound that scales as $O(\gamma \log(T) +K)$. Hence OSUB offers better performance guarantees than LSE as soon as the time horizon satisfies $T \ge \exp(K/\gamma D)$. Although this is not explicitly mentioned in \cite{yu2011}, we believe that LSE was meant to address bandits where the number of arms is not negligible compared to the time horizon.

(c) We further investigate OSUB performance in non-stationary environments where the expected rewards smoothly evolve over time but keep their unimodal structure. 

(d) We conduct numerical experiments and show that \ouralgo significantly outperforms LSE and other classical bandit algorithms when the number of arms is much smaller than the time horizon. 

(e) Finally, we briefly discuss systems with a continuous set of arms. We show that using a simple discretization of the set of arms, UCB-like algorithms are order-optimal, and actually outperform more advanced algorithms such as those proposed in \cite{yu2011}. This result suggests that in discrete unimodal bandits with a very large number of arms, it is wise to first prune the set of arms, so as to reduce its size to a number of the order of $\sqrt{T}/\log(T)$. 

\section{Related work}

Unimodal bandits have received relatively little attention in the literature. They are specific instances of bandits in metric spaces \cite{Kleinberg2004, kleinberg2008, bubeck08}. In this paper, we add unimodality and show how this structure can be optimally exploited. Unimodal bandits have been specifically addressed in \cite{cope09, yu2011}. In \cite{cope09}, bandits with a continuous set of arms are studied, and the author shows that the Kiefer-Wolfowitz stochastic approximation algorithm achieves a regret of the order of $O(\sqrt{T})$ under some strong regularity assumptions on the reward function. In \cite{yu2011}, for the same problem, the authors present LSE, an algorithm whose regret scales as $O(\sqrt{T}\log(T))$ without the need for a strong regularity assumption. The LSE algorithm is based on Kiefer's golden section search algorithm. It iteratively eliminates subsets of arms based on PAC-bounds derived after appropriate sampling. By design, under LSE, the sequence of parameters used for the PAC bounds is pre-defined, and in particular does not depend of the observed rewards. As a consequence, LSE may explore too much sub-optimal parts of the set of arms. For bandits with a continuum set of arms, we actually show that combining an appropriate discretization of the decision space (i.e., reducing the number of arms to $\sqrt{T}/\log(T)$ arms) and the UCB algorithm can outperform LSE in practice (this is due to the adaptive nature of UCB). Note that the parameters used in LSE to get a regret of the order $O(\sqrt{T}\log(T))$ depend on the time horizon $T$.

In \cite{yu2011}, the authors also present an extension of the LSE algorithm to problems with discrete sets of arms, and provide regret upper bounds of this algorithm. These bounds depends on the structure of the graph defining unimodal structure, and on the number of arms as mentioned previously. LSE performs better than classical bandit algorithms only when the number of arms is very large, and actually becomes comparable to the time horizon. Here we are interested in bandits with relatively small number of arms. 

Non-stationary bandits have been studied in \cite{hartland06,garivier08, slivkins08, yu2011}. Except for \cite{slivkins08}, these papers deal with environments where the expected rewards and the best arm change abruptly. This ensures that arms are always well separated, and in turn, simplifies the analysis. In \cite{slivkins08}, the expected rewards evolve according to independent brownian motions. We consider a different, but more general class of dynamic environments: here the rewards smoothly evolve over time. The challenge for such environments stems from the fact  that, at some time instants, arms can have expected rewards arbitrarily close to each other.

Finally, we should mention that bandit problems with structural properties such as those we address here can often be seen as specific instances of problems in the control of Markov chains, see \cite{graves1997}. We leverage this observation to derive regret lower bounds. However, algorithms developed for the control of generic Markov chains are often too complex to implement in practice. Our algorithm, OSUB, is optimal and straightforward to implement.

\section{Model and Objectives}\label{sec:model}

We consider a stochastic multi-armed bandit problem with $K \geq 2$ arms. We discuss problems where the set of arms is continuous in Section \ref{sec:continuous}. Time proceeds in rounds indexed by $n=1,2,\ldots$. Let $X_k(n)$ be the reward obtained at time $n$ if arm $k$ is selected. For any $k$, the sequence of rewards $(X_{k}(n))_{n\ge 1}$ is i.i.d. with distribution and expectation denoted by $\nu_k$ and $\mu_k$ respectively. Rewards are independent across arms. Let $\mu=(\mu_1,\ldots,\mu_K)$ represent the expected rewards of the various arms. At each round, a decision rule or algorithm selects an arm depending on the arms chosen in earlier rounds and their observed rewards. We denote by $k^\pi(n)$ the arm selected under $\pi$ in round $n$. The set $\Pi$ of all possible decision rules consists of policies $\pi$ satisfying: for any $n\ge 1$, if ${\cal F}_n^\pi$ is the $\sigma$-algebra generated by $(k^\pi(t),X_{k^\pi(t)}(t))_{1\le t\le n}$, then $k^\pi(n+1)$ is ${\cal F}_{n}^\pi$-measurable. 

\subsection{Unimodal Structure}

The expected rewards exhibit a {\it unimodal} structure, similar to that considered in \cite{yu2011}. More precisely, there exists an undirected graph $G= (V,E)$ whose vertices correspond to arms, i.e., $V = \{1,\dots,K\}$, and whose edges characterize a partial order (initially unknown to the decision maker) among expected rewards. We assume that there exists a unique arm $k^\star$ with maximum expected reward $\mu^\star$, and that from any sub-optimal arm $k\neq k^\star$, there exists a path $p=(k_1=k,\ldots,k_m=k^\star)$ of length $m$ (depending on $k$) such that for all $i=1,\ldots,m-1$, $(k_i,k_{i+1})\in E$ and $\mu_{k_i} < \mu_{k_{i+1}}$. We denote by ${\cal U}_G$ the set of vectors $\mu$ satisfying this unimodal structure.

This notion of unimodality is quite general, and includes, as a special case, classical unimodality (where $G$ is just a line). Note that we assume that the decision maker knows the graph $G$, but ignores the best arm, and hence the partial order induced by the edges of $G$. 
	
\subsection{Stationary and non-stationary environments}

The model presented above concerns stationary environments, where the expected rewards for the various arms do not evolve over time. In this paper, we also consider non-stationary environments where these expected rewards could evolve over time according to some deterministic dynamics. In such scenarios, we denote by $\mu_k(n)$ the expected reward of arm $k$ at time $n$, i.e., $\mathbb{E}[X_k(n)]=\mu_k(n)$, and $(X_k(n))_{n\ge 1}$ constitutes a sequence of independent random variables with evolving mean. In non-stationary environments, the sequences of rewards are still assumed to be independent across arms. Moreover, at any time $n$, $\mu(n)=(\mu_1(n),\ldots \mu_K(n))$ is unimodal with respect to some fixed graph $G$, i.e., $\mu(n)\in {\cal U}_G$ (note however that the partial order satisfied by the expected rewards may evolve over time). 

\subsection{Regrets}

The performance of an algorithm $\pi\in \Pi$ is characterized by its {\it regret} up to time $T$ (where $T$ is typically large). The way regret is defined differs depending on the type of environment.

{\it Stationary Environments.} In such environments, the regret $R^\pi(T)$ of algorithm $\pi\in \Pi$ is simply defined through the number of times $t_k^\pi(T) = \sum_{1 \leq n \leq T} \indic \{k^\pi(n) = k \}$ that arm $k$ has been selected up to time $T$:
$
R^{\pi}(T) = \sum_{k=1}^K (\mu^{\star} - \mu_k) \EE[ t_k^\pi(T) ].
$
Our objectives are (1) to identify an asymptotic (when $T\to\infty$) regret lower bound satisfied by {\it any} algorithm in $\Pi$, and (2) to devise an algorithm that achieves this lower bound.

{\it Non-stationary Environments.} In such environments, the regret of an algorithm $\pi\in \Pi$ quantifies	 how well $\pi$ tracks the best arm over time.  Let $k^\star(n)$ denote the optimal arm with expected reward $\mu^\star(n)$ at time $n$. The regret of $\pi$ up to time $T$ is hence defined as:
$
R^{\pi}(T) = \sum_{n=1}^T \left( \mu^\star(n) - \EE[\mu_{k^\pi(n)}(n)]\right).
$

\section{Stationary environments}\label{sec:stationary}

In this section, we consider unimodal bandit problems in stationary environments. We derive an asymptotic lower bound of regret when the reward distributions belong to a parametrized family of distributions, and propose OSUB, an algorithm whose regret matches this lower bound.

\subsection{Lower bound on regret}

To simplify the presentation, we assume here that the reward distributions belong to a parametrized family of distributions. More precisely, we define a set of distributions ${\cal V} = \{\nu(\theta) \}_{\theta \in [0,1]}$ parametrized by $\theta\in [0,1]$. The expectation of $\nu(\theta)$ is denoted by $\mu(\theta)$ for any $\theta\in [0,1]$. $\nu(\theta)$ is absolutely continuous with respect to some positive measure $m$ on $\RR$, and we denote by $p(x,\theta)$ its density. The Kullback-Leibler (KL) divergence number between $\nu(\theta)$ and $\nu(\theta')$ is:
$
KL(\theta,\theta^{\prime}) = \int_{\RR}  \log(  p(x,\theta) / p(x,\theta^{\prime}) ) p(x,\theta) m(dx).
$
We denote by $\theta^\star$ a parameter (it might not be unique) such that $\mu(\theta^\star)=\mu^\star$, and we define the minimal divergence number between $\nu(\theta)$ and $\nu(\theta^\star)$ as:
$
I_{\min}(\theta, \theta^{\star}) = \inf_{ \theta\in [0,1]: \mu( \theta^{\prime} ) \geq \mu^{\star}} KL(\theta,\theta^{\prime}).
$

Finally, we say that arm $k$ has parameter $\theta_k$ if $\nu_k =  \nu(\theta_k)$, and we denote by $\Theta_G$ the set of all parameters $\theta=(\theta_1, \dots, \theta_K)\in [0,1]^K$ such that the corresponding expected rewards are unimodal with respect to graph $G$: $\mu=(\mu_1,\ldots,\mu_K)\in {\cal U}_G$. Of particular interest is the family of Bernoulli distributions: the support of $m$ is $\{0,1\}$, $\mu(\theta)=\theta$, and $I_{\min}(\theta,\theta^\star) = I(\theta,\theta^\star)$ where $I(\theta,\theta^\star)=\theta\log({\theta\over\theta^\star})+(1-\theta)\log({1-\theta\over 1-\theta^\star})$ is KL divergence number between Bernoulli distributions of respective means $\theta$ and $\theta^\star$.

We are now ready to derive an asymptotic regret lower in parametrized unimodal bandit problems as defined above. Without loss of generality, we restrict our attention to so-called uniformly good algorithms, as defined in \cite{lai1985} (uniformly good algorithms exist as shown later on). We say that $\pi\in\Pi$ is uniformly good if for all $\theta \in \Theta_G$, we have that $R^{\pi}(T) = o(T^{a})$ for all $a > 0$. 

\begin{theorem}\label{th:graves_lai}
Let $\pi\in \Pi$ be a uniformly good algorithm, and assume that $\nu_k = \nu(\theta_k) \in {\cal V}$ for all $k$. Then for any $\theta\in \Theta_G$,
\eq{
\lim \inf_{T \to +\infty} \frac{R^{\pi}(T)}{ \log(T)}  \geq c(\theta) = \sum_{(k,k^*) \in E} \frac{ \mu^{\star} - \mu_k}{I_{\min}(\theta_k, \theta^{\star})}.
}
\end{theorem}

The above theorem is a consequence of results in optimal control of Markov chains \cite{graves1997}. All proofs are presented in appendix. As in classical discrete bandit problems, the regret scales at least logarithmically with time (the regret lower bound derived in \cite{lai1985} is obtained from Theorem \ref{th:graves_lai} assuming that $G$ is the complete graph). We also observe that the unimodal structure, if optimally exploited, can bring significant performance improvements: the regret lower bound does not depend on the size $K$ of the decision space. Indeed $c(\theta)$ includes only terms corresponding to arms that are neighbors in $G$ of the optimal arm (as if one could learn without regret that all other arms are sub-optimal). 

In the case of Bernoulli rewards, the lower regret bound becomes $\log(T)\sum_{(k,k^*) \in E} \frac{ \mu^{\star} - \mu_k}{I(\theta_k, \theta^{\star})}$. Note that LSE and GLSE, the algorithms proposed in \cite{yu2011}, have performance guarantees that do not match our lower bound: when $G$ is a line, LSE achieves a regret bounded by $41/\Delta^2\log(T)$, whereas in the general case, GLSE incurs a regret of the order of $O(\gamma D\log(T))$ where $\gamma$ is the maximal degree of vertices in $G$, and $D$ is its diameter. The performance of LSE critically depends on the graph structure, and the number of arms. Hence there is an important gap between the performance of existing algorithms and the lower bound derived in Theorem \ref{th:graves_lai}. In the next section, we close this gap and propose an asymptotically optimal algorithm.

\subsection{The OSUB Algorithm}

We now describe OSUB, a simple algorithm whose regret matches the lower bound derived in Theorem of~\ref{th:graves_lai} for Bernoulli rewards, i.e., OSUB is asymptotically optimal. The algorithm is based on KL-UCB proposed in \cite{lai1987, cappe2012}, and uses KL-divergence upper confidence bounds to define an {\it index} for each arm. OSUB can be readily extended to systems where reward distributions are within one-parameter exponential families by simply modifying the definition of arm indices as done in \cite{cappe2012}. In OSUB, each arm is attached an index that resembles the KL-UCB index, but the arm selected at a given time is the arm with maximal index within the neighborhood in $G$ of the arm that yielded the highest empirical reward. Note that since the sequential choices of arms are restricted to some neighborhoods in the graph, OSUB is not an index policy. To formally describe OSUB, we need the following notation. For $p \in [0,1]$, $s \in \NN$, and $n \in \NN$, we define:
\begin{align}
F(p,s,n) =& \sup \{ q \geq p : \nonumber\\
& sI(p,q) \leq \log(n) + c \log( \log (n) ) \},
\end{align}
with the convention that $F(p,0,n) = 1$, and $F(1,s,n) = 1$, and where $c>0$ is a constant. Let $k(n)$ be the arm selected under OSUB at time $n$, and let $t_k(n)$ denote the number of times arm $k$ has been selected up to time $n$. The empirical reward of arm $k$ at time $n$ is $\hat \mu_k(n) = {1\over t_k(n)} \sum_{t=1}^{n} \indic \{ k(t)=k  \} X_k(t)$, if $t_k(n) > 0$ and $\hat \mu_k(n) = 0$ otherwise. We denote by $L(n) = \arg \max_{1 \leq k \leq K} \hat \mu_k(n)$ the index of the arm with the highest empirical reward (ties are broken arbitrarily). Arm $L(n)$ is referred to as the {\it leader} at time $n$. Further define $l_k(n) = \sum_{t=1}^n  \indic \{ L(t) = k  \}$ the number of times arm $k$ has been the leader up to time $n$. Now the index of arm $k$ at time $n$ is defined as: 
\eqs{
b_k(n) = F(\hat \mu_k(n),t_k(n),l_k( L(n)) ).
}
Finally for any $k$, let $N(k)=\{k':(k',k)\in E\}\cup\{k\}$ be the neighborhood of $k$ in $G$. 
The pseudo-code of OSUB is presented below.

\begin{separation}
\vspace{-0.2cm}
    {\bf Algorithm} OSUB 
\vspace{-0.6cm}\separator
\vspace{-0.4cm}
Input: graph $G=(V,E)$\\
For $n \ge 1$, select the arm $k(n) $ where:
$$ k(n) = \begin{cases}  L(n)  & \text{if } {l_{L(n)}(n) - 1 \over\gamma+1} \in \NN, \\
	  \displaystyle \arg\max_{k\in N(L(n))} b_k(n) & \text{otherwise,}
	  \end{cases} $$
where $\gamma$ is the maximal degree of nodes in $G$ and ties are broken arbitrarily.
\vspace{-0.2cm}
\end{separation}
Note that OSUB forces us to select the current leader often: $L(n)$ is chosen when $l_{L(n)}(n)-1$ is a multiple of $\gamma+1$. This ensures that the number of times an arm has been selected is at least proportional to the number of times this arm has been the leader. This property significantly simplifies the regret analysis, but it could be removed.

\subsection{Finite-time analysis of OSUB}

Next we provide a finite time analysis of the regret achieved under OSUB. Let $\Delta$ denote the minimal separation between an arm and its best adjacent arm: $\Delta = \min_{1 \leq k \leq K} \max_{k^\prime: (k,k^\prime) \in E} \mu_{k^\prime} - \mu_{k}$. Note that $\Delta$ is not known a priori.

\begin{theorem}\label{th:kluucb_finite}
Assume that the rewards lie in [0,1] (i.e., the support of $\nu_k$ is included in $[0,1]$, for all $k$), and that $(\mu_1,\ldots,\mu_K)\in {\cal U}_G$. The number of times suboptimal arm $k$ is selected under OSUB satisfies: for all $\epsilon>0$ and all $T\ge 3$,
\als{
\EE[ t_k(T) ] &\leq   \begin{cases} (1 + \epsilon) \frac{ \log(T) + c \log(\log(T))}{ I(\mu_k,\mu^*)}  
 & \text{ if }  (k,k^{\star}) \in E, \\
\;\;\;\; +  C_1 \log \log (T) + \frac{C_2}{T^{\beta(\epsilon)}}  & \\ 
\frac{C_3}{\Delta^2} & \text{ otherwise,}  \end{cases} 
}
where $\beta(\epsilon) > 0$, and $0<C_1 < 7$, $C_2>0$, $C_3>0$ are constants.
\end{theorem}

To prove this upper bound, we analyze the regret accumulated (i) when the best arm $k^\star$ is the leader, and (ii) when the leader is arm $k\neq k^\star$. (i) When $k^\star$ is the leader, the algorithm behaves like KL-UCB restricted to the arms around $k^\star$, and the regret at these rounds can be analyzed as in \cite{cappe2012}. (ii) Bounding the number of rounds where $k\neq k^\star$ is not the leader is more involved. To do this, we decompose this set of rounds into further subsets (such as the time instants where $k$ is the leader and its mean is not well estimated), and control their expected cardinalities using concentration inequalities. Along the way, we establish Lemma~\ref{lem:concentr}, a new concentration inequality of independent interest. 

\begin{lemma}\label{lem:concentr}
Let $\{ Z_t \}_{t \in \mathbb{Z}}$ be a sequence of independent random variables with values in $[0,B]$. Define ${\cal F}_n$ the $\sigma$-algebra generated by $\{ Z_t \}_{t \leq n}$ and the filtration ${\cal F} = ( {\cal F}_n )_{n \in \mathbb{Z}} $. Consider $s \in \NN$, $n_0 \in \mathbb{Z}$ and $T \geq n_0$. We define $S_n = \sum_{t=n_0}^n B_t (Z_t - \EE[Z_t])$, where $B_t \in \{0,1\}$ is a ${\cal F}_{t-1}$-measurable random variable. Further define $t_n = \sum_{t=n_0}^n B_t$. Define $\phi \in \{n_0,\dots,T+1\}$ a ${\cal F}$-stopping time such that either $t_{\phi} \geq s$ or $\phi = T+1$. 

 Then we have that:
 $
 \PP[ S_{\phi}  \geq t_{\phi} \delta \;,\;  \phi \leq T  ] \leq \exp( -2 s \delta^2 B^{-2}).
 $
 As a consequence:
 $
 \PP[ | S_{\phi} | \geq t_{\phi} \delta \;,\;  \phi \leq T  ] \leq 2 \exp( -2 s \delta^2 B^{-2}).
 $
\end{lemma}
Lemma~\ref{lem:concentr} concerns the sum of products of i.i.d. random variables and of a previsible sequence, evaluated at a stopping time (for the natural filtration). We believe that concentration results for such sums can be instrumental in bandit problems, where typically, we need information about the empirical rewards at some specific random time epochs (that often are stopping times). Refer to the appendix for a  proof. A direct consequence of Theorem \ref{th:kluucb_finite} is the asymptotic optimality of OSUB in the case of Bernoulli rewards:

\begin{corollary}
Assume that rewards distributions are Bernoulli (i.e for any $k$, $\nu_k \sim \text{Bernoulli}(\theta_k)$), and that $\theta\in \Theta_G$. Then the regret achieved under $\pi$=OSUB satisfies:
$
\lim \sup_{T \to +\infty} R^{\pi}(T) / \log(T) \leq c(\theta).
$
\end{corollary}


\section{Non-stationary environments}\label{sec:non_stationary}

We now consider time-varying environments. We assume that the expected reward of each arm varies smoothly over time, i.e., it is Lipschitz continuous: for all $n, n' \ge 1$ and $1 \leq k \leq K$: $|\mu_k(n) - \mu_k(n^\prime)  | \leq \sigma |n - n^\prime|$.

We further assume that the unimodal structure is preserved (with respect to the same graph $G$): for all $n\ge 1$, $\mu(n)\in {\cal U}_G$. Considering smoothly varying rewards is more challenging than scenarios where the environment is abruptly changing. The difficulty stems from the fact that the rewards of two or more arms may become arbitrarily close to each other (this happens each time the optimal arm changes), and in such situations, regret is difficult to control. To get a chance to design an algorithm that efficiently tracks the best arm, we need to make some assumption to limit the proportion of time when the separation of arms becomes too small. Define for $T \in \NN$, and $\Delta > 0$:
\eqs{
H(\Delta,T) = \sum_{n=1}^{T} \sum_{ (k,k^\prime) \in E } \indic \{ | \mu_k(n) - \mu_{k^\prime}(n) | < \Delta  \}.
}
\begin{assumption}\label{ass:smooth}
There exists a function $\Phi$ and $\Delta_0$ such that for all $\Delta < \Delta_0$:
$
\lim \sup_{T \to +\infty} H(\Delta,T)/T \leq \Phi(K) \Delta.
$
\end{assumption}

\subsection{\ouralgo with a Sliding Window}
	
To cope with the changing environment, we modify the \ouralgo algorithm, so that decisions are based on past choices and observations over a time-window of fixed duration equal to $\tau+1$ rounds. The idea of adding a sliding window to algorithms initially designed for stationary environments is not novel \cite{garivier08}; but here, the unimodal structure and the smooth evolution of rewards make the regret analysis more challenging.

Define: $t^{\tau}_k(n) = \sum_{t =n-\tau}^{n} \indic\{ k(t) = k \}$; $\hat\mu^{\tau}_k(n) = (1/t^{\tau}_k(n)) \sum_{t =n-\tau}^{n} \indic\{ k(t) = k \} X_k(t)$ if $t^{\tau}_k(n) > 0$ and $\hat\mu^{\tau}_k(n) = 0$ otherwise; $L^{\tau}(n) = \arg \max_{1 \leq k \leq K} \hat\mu^{\tau}_k(n)$;          $l^{\tau}_k(n) = \sum_{t=n-\tau}^{n} \indic\{ L^{\tau}(t) = k \}$. The index of arm $k$ at time $n$ then becomes: 
$
b^{\tau}_k(n) = F(\hat\mu^{\tau}_k(n), t_k^\tau(n), l_k^\tau(L^\tau(n))).
$
The pseudo-code of \ouralgosw is presented below.
\begin{separation}
\vspace{-0.2cm}
    {\bf Algorithm} \ouralgosw
\vspace{-0.6cm}\separator
\vspace{-0.4cm}
Input: graph $G=(V,E)$, window size $\tau+1$\\
For $n \ge 1$, select the arm $k(n) $ where:
$$ k(n) = \begin{cases}  L^\tau(n)  & \text{if } {l_{L^\tau(n)}^\tau(n) - 1\over\gamma+1} \in \NN, \\
	  \displaystyle \arg\max_{k\in N(L^\tau(n))} b_k^\tau(n) & \text{otherwise.}
	  \end{cases} $$
\vspace{-0.2cm}
\end{separation}

\subsection{Regret Analysis}

In non-stationary environments, achieving sublinear regrets is often not possible. In \cite{garivier08}, the environment is subject to abrupt changes or breakpoints. It is shown that if the density of breakpoints is strictly positive, which typically holds in practice, then the regret of any algorithm has to scale linearly with time. We are interested in similar scenarios, and consider smoothly varying environments where the number of times the optimal arm changes has a positive density. The next theorem provides an upper bound of the regret per unit of time achieved under SW-OSUB. This bound holds for any non-stationary environment with $\sigma$-Lipschitz rewards. 

\begin{theorem}	\label{th:nonstat}
Let $\Delta$: $2\tau\sigma < \Delta < \Delta_0$. Assume that for any $n\ge 1$, $\mu(n)\in {\cal U}_G$ and $\mu^\star(n)\in [a,1-a]$ for some $a>0$. Further suppose that $\mu_k(\cdot)$ is $\sigma$-Lipschitz for any $k$. The regret per unit time under $\pi=$\ouralgosw with a sliding window of size $\tau+1$ satisfies: if $a>\sigma\tau$, then for any $T\ge 1$,
\als{	
\frac{R^{\pi}(T)}{T} &\leq  \frac{H(\Delta,T)}{T}(1 + \Delta)  +  \frac{C_1 K \log(\tau)}{\tau (\Delta - 4 \tau \sigma)^2} \sk
 & +\gamma \Lp 1 + g_0^{-1/2} \Rp \frac{\log(\tau) + c \log(\log(\tau)) + C_2}{2 \tau (\Delta - 2 \tau \sigma)^2} ,
}
where $C_1,C_2$ are positive constants and $g_0=(a - \sigma\tau)(1-a + \sigma\tau)/2$.
\end{theorem}

\begin{corollary} Assume that for any $n\ge 1$, $\mu(n)\in {\cal U}_G$ and $\mu^\star(n)\in [a,1-a]$ for some $a>0$, and that $\mu_k(\cdot)$ is $\sigma$-Lipschitz for any $k$. Set $\tau = \sigma^{-3/4} \log(1/\sigma) / 8 $. The regret per unit of time of $\pi=$\ouralgosw with window size $\tau+1$ satisfies: 
$$
\lim \sup_{T\to\infty} \frac{R^{\pi}(T)}{T}  \leq C \Phi(K) \sigma^{\frac{1}{4}} \log \Lp \frac{1}{\sigma} \Rp ( 1 + K j(\sigma) ),
$$
for some constant $C>0$, and some function $j$ such that $\lim_{\sigma\to 0^+} j(\sigma)  =0.$
\end{corollary}

These results state that the regret per unit of time achieved under \ouralgosw decreases and actually vanishes when the speed at which expected rewards evolve decreases to 0. Also observe that the dependence of this regret bound in the number of arms is typically mild (in many practical scenarios, $\Phi(K)$ may actually not depend on $K$).

The proof of Theorem \ref{th:nonstat} relies on the same types of arguments as those used in stationary environments. To establish the regret upper bound, we need to evaluate the performance of the KL-UCB algorithm in non-stationary environments (the result and the corresponding analysis are presented in appendix).

\section{Continuous Set of Arms}\label{sec:continuous}
 
In this section, we briefly discuss the case where the decision space is continuous. The set of arms is $[0,1]$, and the expected reward function $\mu:[0,1] \to \RR$ is assumed to be Lipschitz continuous, and unimodal: there exists $x^\star\in [0,1]$ such that $\mu(x^\prime) \geq \mu(x)$ if $x^\prime \in [x ,x^{\star}]$ or $x^\prime \in [x^{\star} , x]$. Let $\mu^\star=\mu(x^\star)$ denote the highest expected reward. A decision rule selects at any round $n\ge 1$ an arm $x$ and observes the corresponding reward $X(x,n)$. For any $x\in [0,1]$, $(X(x,n))_{n\ge 1}$ is an i.i.d. sequence. We make the following additional assumption on function $\mu$.

\begin{assumption}\label{ass:22}
There exists $\delta_0>0$ such that (i) for all $x , y$ in $[x^\star,x^\star+\delta_0]$ (or in $[x^\star-\delta_0,x^\star]$), $C_1 |x-y|^\alpha \leq |\mu(x) - \mu(y)  |$; (ii) for $\delta \leq \delta_0$, if $|x - x^*| \leq \delta$, then $| \mu(x^*) - \mu(x) | \leq C_2 \delta^\alpha$. 
\end{assumption}

This assumption is more general than that used in \cite{yu2011}. In particular it holds for functions with a {\it plateau} and a {\it peak}: $\mu(x)=\max(1-|x-x^\star|/\epsilon,0)$. Now as for the case of a discrete set of arms, we denote by $\Pi$ the set of possible decision rules, and the regret achieved under rule $\pi\in \Pi$ up to time $T$ is:
$
R^{\pi}(T) = T \mu^{\star} - \sum_{n=1}^T  \mathbb{E}[\mu(x^\pi(n))],
$
where $x^\pi(n)$ is the arm selected under $\pi$ at time $n$. 

There is no known precise asymptotic lower bound for continuous bandits. However, we know that for our problem, the regret must be at least of the order of $O(\sqrt{T})$ up to logarithmic factor. In \cite{yu2011}, the authors show that the LSE algorithm achieves a regret scaling as $O(\sqrt{T}\log(T))$, under more restrictive assumptions. We show that combining discretization and the UCB algorithm as initially proposed in \cite{Kleinberg2004} yields lower regrets than LSE in practice (see Section \ref{sec:numerical}), and is order-optimal, i.e., the regret grows as $O(\sqrt{T}\log(T))$.  

For $\delta > 0$, we define a discrete bandit problem with $K = \ceil{1/\delta}$ arms, and where the rewards of $k$-th arm are distributed as $X((k-1)/\delta,n)$. The expected reward of the $k$-th arm is $\mu_k = \mu((k-1)/\delta)$. Let $\pi$ be an algorithm running on this discrete bandit problem. The regret of $\pi$ for the initial continuous bandit problem is at time $T$:\\
$ 
R^{\pi}(T) = T \mu^{\star} - \sum_{k=1}^{ \ceil{1/\delta} } \mu_k \EE[t_k^\pi(T)].
$
We denote by UCB($\delta$) the UCB algorithm \cite{auer2002} applied to the discretized bandit. In the following proposition, we show that when $\delta= (\log(T)/\sqrt{T} )^{1/\alpha}$, UCB($\delta$) is order-optimal. In practice, one may not know the time horizon $T$ in advance. In this case, using the ``doubling trick'' (see e.g. \cite{CesaBianchi06})  would incur an additional logarithmic multiplicative factor in the regret.
	
\begin{proposition}\label{th:unimodal_cont}
Consider a unimodal bandit on $[0,1]$ with rewards in $[0,1]$ and satisfying Assumption \ref{ass:22}. Set $\delta= (\log(T)/\sqrt{T} )^{1/\alpha}$. The regret under UCB($\delta$) satisfies:
\eqs{ \lim \sup_{T\to\infty} {R^{\pi}(T) \over \sqrt{T} \log(T)} \leq C_2 3^{\alpha} + 16 /C_1.}
\end{proposition}

\section{Numerical experiments}\label{sec:numerical}

\subsection{Discrete bandits}

We compare the performance of our algorithm to that of KL-UCB \cite{cappe2012}, LSE \cite{yu2011}, UCB \cite{auer2002}, and UCB-U. The latter algorithm is obtained by applying UCB restricted to the arms which are adjacent to the current leader as in OSUB. We add the prefix "SW" to refer to Sliding Window versions of these algorithms.

\begin{figure}
\begin{center}
	\includegraphics[width=0.7\columnwidth]{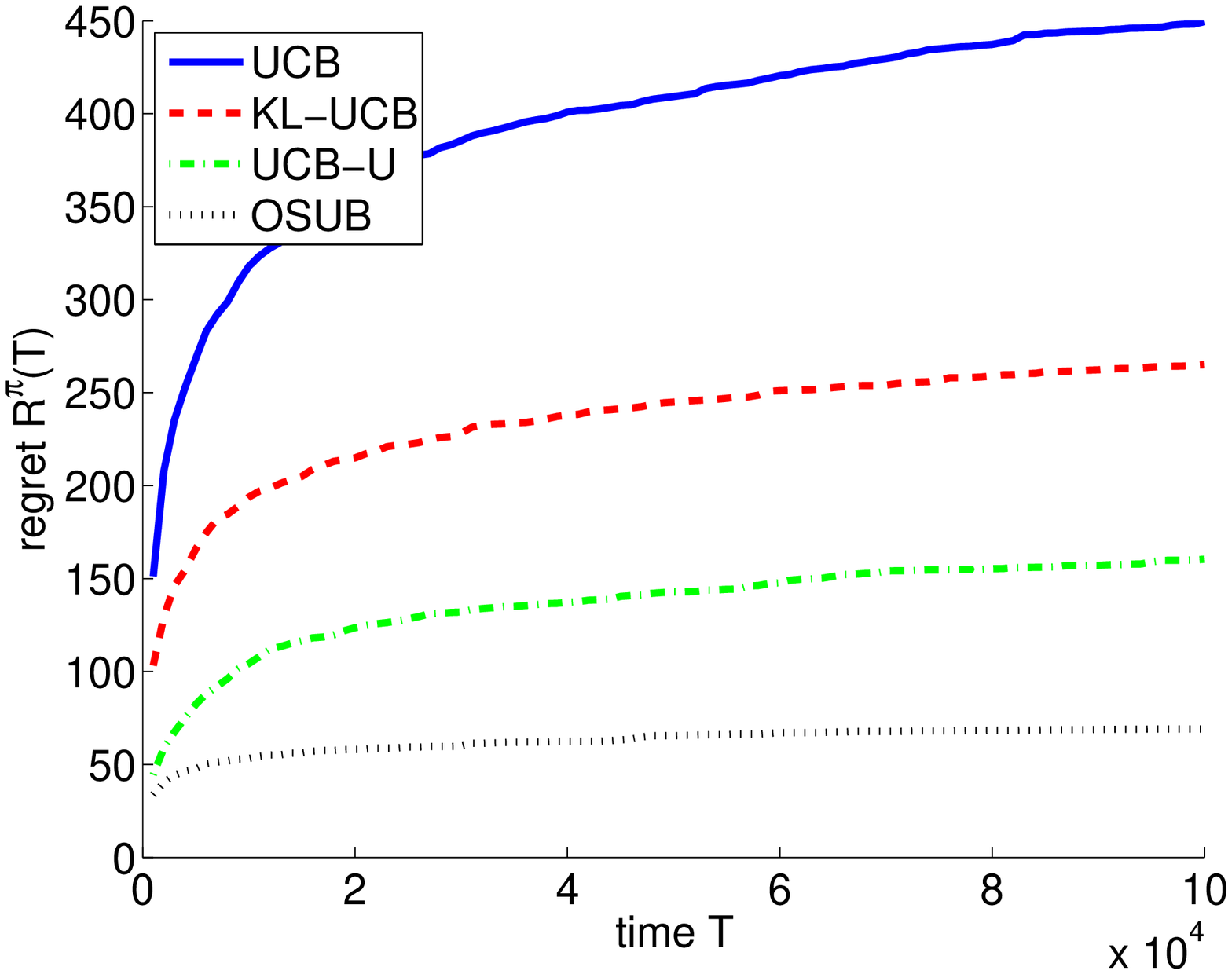}
	\caption{Regret vs. time in stationary environments -- $K=17$ arms.}
	\label{fig:compare_stationary}
\end{center}
\end{figure}

{\it Stationary environments.} In our first experiment, we consider $K=17$ arms with Bernoulli rewards of respective averages $\mu=(0.1, 0.2, .... , 0.9 , 0.8,  \dots, 0.1)$. The rewards are unimodal (the graph $G$ is simply a line). The regret achieved under the various algorithms is presented in Figure~\ref{fig:compare_stationary} and Table~\ref{table:regret_log}. The parameters in LSE algorithm are chosen as suggested in Proposition 4.5 \cite{yu2011}. Regrets are calculated averaging over $50$ independent runs. \ouralgo significantly outperforms all other algorithms. The regret achieved under LSE is not presented in Figure~\ref{fig:compare_stationary}, because it is typically much larger than that of other algorithms. This poor performance can be explained by the non-adaptive nature of LSE, as already discussed earlier. LSE can beat UCB when the number of arms is not negligible compared to the time horizon (e.g. in Figure 4 in \cite{yu2011}, $K=250.000$ and the time horizon is less than $3K$): in such scenarios, UCB-like algorithms perform poorly because of their initialization phase (all arms have to be tested once).

\begin{figure}
\begin{center}
	\includegraphics[width=0.7\columnwidth]{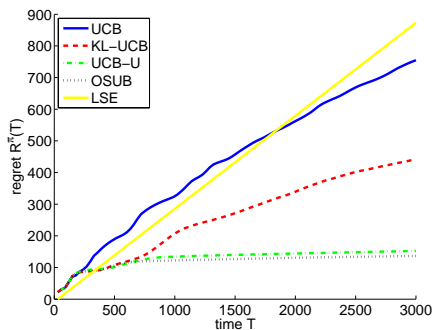}
	\caption{Regret vs. time in stationary environments -- $K=129$ arms.}
	\label{fig:stat2}
\end{center}
\end{figure}

In Figure \ref{fig:stat2}, the number of arms is 129, and the expected rewards form a triangular shape as in the previous example, with minimum and maximum equal to 0.1 and 0.9, respectively. Similar observations as in the case of 17 arms can be made. We deliberately restrict the plot to small time horizons: this corresponds to scenarios where LSE can perform well.

\begin{table}
\begin{center}
\begin{tabular}{llll}
\hline
 T  & 1000 & 10000 & 100000\\
\hline
UCB & 30.1 & 35.1 & 39\\
KL-UCB & 18.8 & 21.4 & 23\\
UCB-U & 8.5 & 11.7 & 13.9\\
OSUB & 5.8 & 5.9 & 6\\
LSE & 36.3 & 271.5 & 999.1\\
\hline
\end{tabular}
\end{center}
\caption{ $R^{\pi}(T) / \log(T)$ for different algorithms -- 17 arms.}
\label{table:regret_log}
\end{table}

\begin{figure}
\begin{center}
	\includegraphics[width=0.7\columnwidth]{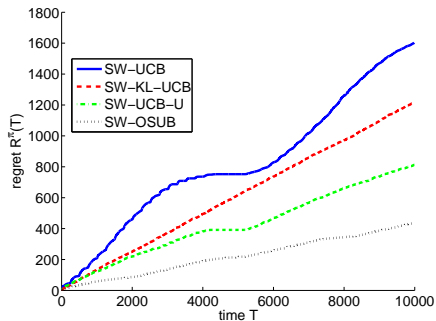}
	\caption{Regret vs. time in a slowly varying environment -- $K=10$ arms, $\sigma = 10^{-3}$.}
	\label{fig:compare_sw_regret_time}
\end{center}
\end{figure}

\begin{figure}
\begin{center}
	\includegraphics[width=0.7\columnwidth]{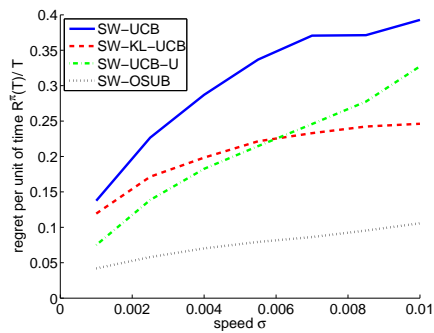}
	\caption{Regret per unit of time $R^{\pi}(T)/T$ vs. speed $\sigma$ -- $K=10$ arms.}
	\label{fig:compare_sw_regret_speed}
\end{center}
\end{figure}

{\it Non-stationary environments.} We now investigate the performance of \ouralgosw in a slowly varying environment. There are $K=10$ arms whose expected rewards form a moving triangle: for $k=1,\ldots,K$, $\mu_k(n) = (K-1)/K - | w(n) - k |/K$, where  $w(n) = 1 + (K - 1)( 1 +  \sin( n \sigma) )/2$. Figure~\ref{fig:compare_sw_regret_time} presents the regret as a function of time under various algorithms when the speed at which the environment evolves is $\sigma= 10^{-3}$. The window size are set as follows for the various algorithms: $\tau =\sigma^{-4/5}$ for SW-UCB and SW-KL-UCB (the rationale for this choice is explained in appendix), $\tau = \sigma^{-3/4} \log(1/\sigma)/8$ for SW-UCB-U and OSUB. In Figure~\ref{fig:compare_sw_regret_speed}, we show how the speed $\sigma$ impacts the regret per time unit. \ouralgosw provides the most efficient way of tracking the optimal arm.

\subsection{Continuous bandits}

\begin{figure}
\begin{center}
	\includegraphics[width=0.7\columnwidth]{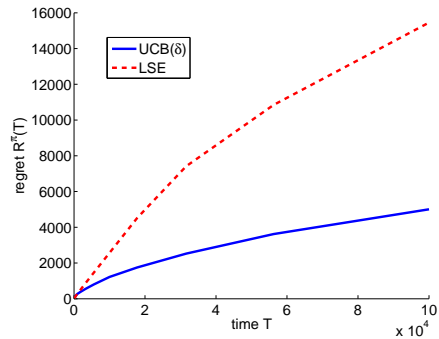}
	\caption{Regret vs. time for a continuous set of arms.}
	\label{fig:compare_continuous}
\end{center}
\end{figure}

In Figure \ref{fig:compare_continuous}, we compare the performance of the LSE and UCB($\delta$) algorithms when the set of arms is continuous. The expected rewards form a triangle: $\mu(x) = 1/2 - |x - 1/2|$ so that $\mu^\star = 1/2$ and $x^\star = 1/2$. The parameters used in LSE are those given in \cite{yu2011}, whereas the discretization parameter $\delta$ in UCB($\delta$) is set to $\delta = \log(T) / \sqrt{T}$. UCB($\delta$) significantly outperforms LSE at any time: an appropriate discretization of continuous bandit problems might actually be more efficient than other methods based on ideas taken from classical optimization theory.
	
\begin{figure}
\begin{center}
	\includegraphics[width=0.7\columnwidth]{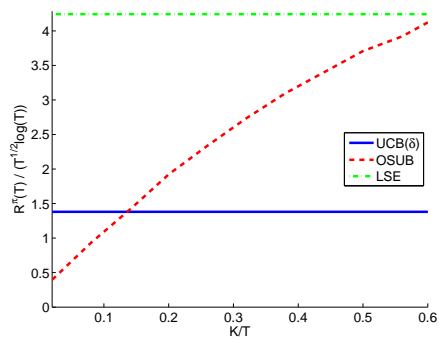}
	\caption{Normalized regret vs. $K/T$, $T = 5.10^{4}$ for a continuous set of arms.}
	\label{fig:number_arms}
\end{center}
\end{figure}

Figure \ref{fig:number_arms} compares the regret of the discrete version of LSE (with optimized parameters), and of OSUB as the number of arms $K$ grows large, $T=50,000$. The average rewards of arms are extracted from the triangle used in the continuous bandit, and we also provide the regret achieved under UCB($\delta$). OSUB outperforms UCB($\delta$) even if the number of arms gets as large as 7500! OSUB also beats LSE unless the number of arms gets bigger than $0.6\times T$. 
	

\section{Conclusion}\label{sec:conclusion}

In this paper, we address stochastic bandit problems with a unimodal structure, and a finite set of arms. We provide asymptotic regret lower bounds for these problems and design an algorithm that asymptotically achieves the lowest regret possible. Hence our algorithm optimally exploits the unimodal structure of the problem. Our preliminary analysis of the continuous version of this bandit problem suggests that when the number of arms become very large and comparable to the time horizon, it might be wiser to prune the set of arms before actually running any algorithm. 

\clearpage
\newpage
\bibliography{RA}

\begin{thebibliography}{23}
\providecommand{\natexlab}[1]{#1}
\providecommand{\url}[1]{\texttt{#1}}
\expandafter\ifx\csname urlstyle\endcsname\relax
  \providecommand{\doi}[1]{doi: #1}\else
  \providecommand{\doi}{doi: \begingroup \urlstyle{rm}\Url}\fi

\bibitem[Agrawal(1995)]{agrawal95}
Agrawal, R.
\newblock The continuum-armed bandit problem.
\newblock \emph{SIAM J. Control and Optimization}, 33\penalty0 (6):\penalty0
  1926--1951, November 1995.

\bibitem[Auer et~al.(2002)Auer, Cesa-Bianchi, and Fischer]{auer2002}
Auer, P., Cesa-Bianchi, N., and Fischer, P.
\newblock Finite time analysis of the multiarmed bandit problem.
\newblock \emph{Machine Learning}, 47\penalty0 (2-3):\penalty0 235--256, 2002.

\bibitem[B.(2005)]{edelman05}
B., Edelman.
\newblock Strategic bidder behavior in sponsored search auctions.
\newblock In \emph{Proc. of {W}orkshop on {S}ponsored {S}earch {A}uctions,
  {ACM} {E}lectronic {C}ommerce}, pp.\  192--198, 2005.

\bibitem[Bubeck \& Cesa-Bianchi(2012)Bubeck and Cesa-Bianchi]{bubeck12}
Bubeck, S. and Cesa-Bianchi, N.
\newblock Regret analysis of stochastic and nonstochastic multi-armed bandit
  problems.
\newblock \emph{Foundations and Trends in Machine Learning}, 5\penalty0
  (1):\penalty0 1--122, 2012.

\bibitem[Bubeck et~al.(2008)Bubeck, Munos, Stoltz, and Szepesv\'ari]{bubeck08}
Bubeck, S., Munos, R., Stoltz, G., and Szepesv\'ari, C.
\newblock Online optimization in x-armed bandits.
\newblock In \emph{Advances in Neural Information Processing Systems 22}, 2008.

\bibitem[Capp\'e et~al.(2013)Capp\'e, Garivier, Maillard, Munos, and
  Stoltz]{cappe2012}
Capp\'e, O., Garivier, A., Maillard, O., Munos, R., and Stoltz, G.
\newblock Kullback-leibler upper confidence bounds for optimal sequential
  allocation.
\newblock \emph{Annals of Statistics}, 41\penalty0 (3):\penalty0 516--541, June
  2013.

\bibitem[Cesa-Bianchi \& Lugosi(2006)Cesa-Bianchi and Lugosi]{CesaBianchi06}
Cesa-Bianchi, N. and Lugosi, G.
\newblock \emph{Prediction, Learning, and Games}.
\newblock Cambridge University Press, 2006.

\bibitem[Cope(2009)]{cope09}
Cope, E.~W.
\newblock Regret and convergence bounds for a class of continuum-armed bandit
  problems.
\newblock \emph{IEEE Trans. Automat. Contr.}, 54\penalty0 (6):\penalty0
  1243--1253, 2009.

\bibitem[Dani et~al.(2008)Dani, Hayes, and Kakade]{dani08}
Dani, V., Hayes, T.~P., and Kakade, S.~M.
\newblock Stochastic linear optimization under bandit feedback.
\newblock In \emph{Proc. of Conference On Learning Theory (COLT)}, pp.\
  355--366, 2008.

\bibitem[Flaxman et~al.(2005)Flaxman, Kalai, and McMahan]{kalai05}
Flaxman, A., Kalai, A.~T., and McMahan, H.~B.
\newblock Online convex optimization in the bandit setting: gradient descent
  without a gradient.
\newblock In \emph{Proc. of {ACM/SIAM} symposium on {D}iscrete {A}lgorithms
  (SODA)}, pp.\  385--394, 2005.

\bibitem[Garivier \& Capp\'e(2011)Garivier and Capp\'e]{garivier2011}
Garivier, A. and Capp\'e, O.
\newblock The {KL-UCB} algorithm for bounded stochastic bandits and beyond.
\newblock In \emph{Proc. of Conference On Learning Theory (COLT)}, 2011.

\bibitem[Garivier \& Moulines(2008)Garivier and Moulines]{garivier08}
Garivier, A. and Moulines, E.
\newblock On upper-confidence bound policies for non-stationary bandit
  problems.
\newblock In \emph{Proc. of Algorithmic Learning Theory (ALT)}, 2008.

\bibitem[Gittins(1989)]{gittins89}
Gittins, J.C.
\newblock \emph{Bandit Processes and Dynamic Allocation Indices}.
\newblock John Wiley, 1989.

\bibitem[Graves \& Lai(1997)Graves and Lai]{graves1997}
Graves, T.~L. and Lai, T.~L.
\newblock Asymptotically efficient adaptive choice of control laws in
  controlled markov chains.
\newblock \emph{SIAM J. Control and Optimization}, 35\penalty0 (3):\penalty0
  715--743, 1997.

\bibitem[Hartland et~al.(2007)Hartland, Baskiotis, Gelly, Teytaud, and
  Sebag]{hartland06}
Hartland, C., Baskiotis, N., Gelly, S., Teytaud, O., and Sebag, M.
\newblock Change point detection and meta-bandits for online learning in
  dynamic environments.
\newblock In \emph{Proc. of conf{\'e}rence francophone sur l'apprentissage
  automatique (CAp07)}, 2007.

\bibitem[Hoeffding(1963)]{Hoeffding1963}
Hoeffding, W.
\newblock Probability inequalities for sums of bounded random variables.
\newblock \emph{Journal of the American Statistical Association}, 58\penalty0
  (301):\penalty0 pp. 13--30, 1963.

\bibitem[Kleinberg et~al.(2008)Kleinberg, Slivkins, and Upfal]{kleinberg2008}
Kleinberg, R., Slivkins, A., and Upfal, E.
\newblock Multi-armed bandits in metric spaces.
\newblock In \emph{Proc. of the 40th annual ACM {S}ymposium on {T}heory of
  {C}omputing (STOC)}, pp.\  681--690, 2008.

\bibitem[Kleinberg(2004)]{Kleinberg2004}
Kleinberg, R.~D.
\newblock Nearly tight bounds for the continuum-armed bandit problem.
\newblock In \emph{Proc. of the conference on Neural Information Processing
  Systems (NIPS)}, 2004.

\bibitem[Lai(1987)]{lai1987}
Lai, T.~L.
\newblock Adaptive treatment allocation and the multi-armed bandit problem.
\newblock \emph{The Annals of Statistics}, 15\penalty0 (3):\penalty0
  1091--1114, 09 1987.

\bibitem[Lai \& Robbins(1985)Lai and Robbins]{lai1985}
Lai, T.L. and Robbins, H.
\newblock Asymptotically efficient adaptive allocation rules.
\newblock \emph{Advances in Applied Mathematics}, 6\penalty0 (1):\penalty0
  4--2, 1985.

\bibitem[Robbins(1952)]{robbins1952}
Robbins, H.
\newblock Some aspects of the sequential design of experiments.
\newblock \emph{Bulletin of the American Mathematical Society}, 58\penalty0
  (5):\penalty0 527--535, 1952.

\bibitem[Slivkins \& Upfal(2008)Slivkins and Upfal]{slivkins08}
Slivkins, A. and Upfal, E.
\newblock Adapting to a changing environment: the brownian restless bandits.
\newblock In \emph{Proc. of Conference On Learning Theory (COLT)}, pp.\
  343--354, 2008.

\bibitem[Yu \& Mannor(2011)Yu and Mannor]{yu2011}
Yu, J. and Mannor, S.
\newblock Unimodal bandits.
\newblock In \emph{Proc. of International Conference on Machine Learning
  (ICML)}, pp.\  41--48, 2011.

\end{thebibliography}
\bibliographystyle{icml2014}

\clearpage
\newpage

\appendix

{ \Large \bf Appendix }

\vspace{0.5cm}

The appendix is organized as follows. In Section~\ref{sec:graveslai}, we prove Theorem~\ref{th:graves_lai}. In Section~\ref{sec:concentration_inequalities}, we state and prove several concentration inequalities which are the cornerstone of our regret analysis of the OSUB algorithm for both stationary and non-stationary environments. In Section~\ref{sec:proof_stationary} we prove Theorem~\ref{th:kluucb_finite}. In Section~\ref{sec:proof_non_stationary}, we prove Theorem~\ref{th:nonstat}. Finally, Section~\ref{sec:app_continuous} is devoted to the proof of Proposition~\ref{th:unimodal_cont}.

\section{Proof of Theorem~\ref{th:graves_lai} }\label{sec:graveslai}

We derive here a regret lower bound for the unimodal bandit problem. To this aim, we apply the techniques used by Graves and Lai \cite{graves1997} to investigate efficient adaptive decision rules in controlled Markov chains. We recall here their general framework. Consider a controlled Markov chain $(X_t)_{t\ge 0}$ on a finite state space ${\cal S}$ with a control set $U$. The transition probabilities given control $u\in U$ are parametrized by $\theta$ taking values in a compact metric space $\Theta$: the probability to move from state $x$ to state $y$ given the control $u$ and the parameter $\theta$ is $p(x,y;u,\theta)$. The parameter $\theta$ is not known. The decision maker is provided with a finite set of stationary control laws $G=\{g_1,\ldots,g_K\}$ where each control law $g_j$ is a mapping from ${\cal S}$ to $U$: when control law $g_j$ is applied in state $x$, the applied control is $u=g_j(x)$. It is assumed that if the decision maker always selects the same control law $g$ the Markov chain is then irreducible with stationary distribution $\pi_\theta^g$. Now the expected reward obtained when applying control $u$ in state $x$ is denoted by $r(x,u)$, so that the expected reward achieved under control law $g$ is: $\mu_\theta(g)=\sum_x r(x,g(x))\pi_\theta^g(x)$. There is an optimal control law given $\theta$ whose expected reward is denoted $\mu_\theta^{\star}\in \arg\max_{g\in G} \mu_\theta(g)$. Now the objective of the decision maker is to sequentially select control laws so as to maximize the expected reward up to a given time horizon $T$. As for MAB problems, the performance of a decision scheme can be quantified through the notion of regret which compares the expected reward to that obtained by always applying the optimal control law.

We now apply the above framework to our unimodal bandit problem, and we consider $\theta \in {\Theta}_G$. The Markov chain has values in ${\cal S}=\RR$. The set of control laws is $G=\{1,\ldots,K\}$. These laws are constant, in the sense that the control applied by control law $k$ does not depend on the state of the Markov chain, and corresponds to selecting arm $k$. The transition probabilities are: $ p(x,y;k,\theta) = p(y,\theta_k)$. Finally, the reward $r(x,k)$ does not depend on the state and is equal to $\mu(\theta_k)$, which is also the expected reward obtained by always using control law $k$. 

We now fix $\theta \in {\Theta}_G$. Define $KL^k(\theta,\lambda)=KL(\theta_k,\lambda_k)$ for any $k$. Further define the set $B(\theta)$ consisting of all {\it bad} parameters $\lambda \in {\Theta}_G$ such that $k^{\star}$ is not optimal under parameter $\lambda$, but which are statistically {\it indistinguishable} from $\theta$:
$$
B(\theta)=\{ \lambda \in {\Theta}_G : \lambda_{k^{\star}} = \theta_{k^{\star}} \textrm{ and }  \max_k  \mu( \lambda_k ) > \mu( \lambda_{k^{\star}} ) \},
$$
$B(\theta)$ can be written as the union of sets $B_k(\theta)$, $(k,k^\star) \in E$ defined as:
$$
B_k(\theta)=\{ \lambda \in B(\theta) :  \mu(\lambda_k) > \mu(\lambda_{k^{\star}}) \}.
$$
We have that $  B(\theta) =  \cup_{(k,k^\star) \in E} B_k(\theta)$, because if $\mu(\lambda_{k^{\star}}) < \max_k \mu(\lambda_{k})$, then there must exist $k$ such that $(k,k^*) \in E$, and $\mu(\lambda_k) > \mu(\lambda_{k^{\star}})$.
By applying Theorem 1 in \cite{graves1997}, we know that $c(\theta)$ is the minimal value of the following LP:
\begin {eqnarray}
\textrm{min }  & \sum_k c_k(\mu( \theta_{k^{\star}} ) - \mu(\theta_k)) \\
\textrm{s.t. } & \inf_{\lambda \in B_k(\theta)} \sum_{l \neq k^{\star}} c_l KL^{l}(\theta,\lambda) \geq 1 , (k,k^\star) \in E \label{eq:con1}\\
& c_k \geq 0, \quad \forall k .
\end {eqnarray}

Next we show that the constraints (\ref{eq:con1}) on the $c_k$'s are equivalent to:
\begin{equation}\label{eq:con2}
\min_{(k,k^\star) \in E} c_{k} I_{min}(\theta_k, \mu(\theta_{k^\star})) \ge 1.
\end{equation}
Consider $k$ fixed with $(k,k^\star) \in E$. We prove that:
\begin{equation}\label{eq:inf}
\inf_{\lambda \in B_k(\theta)} \sum_{l \neq k^{\star}} c_l KL^{l}(\theta,\lambda) = c_k I_{min}(\theta_k,\mu(\theta_{k^\star})).
\end{equation}
This is simply due to the following two observations:
\begin{itemize}
\item Since $ \mu(\lambda_{k}) > \mu(\theta_{k^\star})$ and the KL divergence is positive:
\als{
\sum_{l \neq k^{\star}} c_l KL^{l}(\theta,\lambda) &\ge c_k KL^{k}(\theta,\lambda) \sk
&\ge  c_k I_{min}(\theta_k,\mu(\theta_{k^\star})).
}
\item For $\epsilon >0$, define $\lambda_\epsilon$ as follows: $\mu(\lambda_k) > \mu(\theta_{k^\star})$ and $KL(\theta_k, \lambda_k) \leq I_{min}(\theta_k,\mu(\theta_{k^\star})) + \epsilon$ and $\lambda_l=\theta_l$ for $l \neq k$. By construction, $\lambda_\epsilon\in B_k(\theta)$, and 
$$
\lim_{\epsilon\to 0} \sum_{l \neq k^{\star}} c_l KL^{l}(\theta,\lambda_\epsilon) = c_k I_{min}(\theta_k,\mu(\theta_{k^\star})).
$$
\end{itemize}

From (\ref{eq:inf}), we deduce that constraints (\ref{eq:con1}) are equivalent to (\ref{eq:con2}) (indeed, for $(k,k^\star) \in E$, (\ref{eq:con1}) is equivalent to $c_{k} I_{min}(\theta_k,\mu(\theta_{k^\star})) \ge 1$). With the constraints (\ref{eq:con2}), the optimization problem becomes straightforward to solve, and its solution yields: 
$$
c(\theta) = \sum_{(k,k^{\star}) \in E}  \frac { \mu( \theta_{k^{\star}} ) - \mu( \theta_k) }{  I_{min}(\theta_k,\mu(\theta_{k^\star}))}.
$$
\ep

\section{Concentration inequalities and Preliminaries}\label{sec:concentration_inequalities}

\subsection{Proof of Lemma~\ref{lem:concentr}}

We prove Lemma~\ref{lem:concentr}, a new concentration inequality which extends Hoeffding's inequality, and is used for the regret analysis in subsequent sections. We believe that Lemma~\ref{lem:concentr} could be useful in a variety of bandit problems, where an upper bound on the deviation of the empirical mean \emph{sampled at a stopping time} is needed. An example would be the probability that the empirical reward of the $k$-th arm deviates from its expectation, when it is sampled for the $s$-th time.

\bp
Let $\lambda > 0$, and define $G_n = \exp( \lambda(S_n - \delta t_n)  ) \indic \{n \leq T \}$. We have that:
	\als{
	\PP[& S_{\phi}  \geq t_{\phi} \delta \;,\;  \phi \leq T  ] \sk 
	&= \PP[ \exp( \lambda(S_{\phi}  - \delta t_{\phi} ) ) \indic \{\phi \leq T \} \geq 1] \sk 
	&= \PP[ G_{\phi}  \geq 1] \leq \EE[ G_{\phi} ].
	}
Next we provide an upper bound for $\mathbb{E}[G_{\phi}]$. We define the following quantities:
	\als{ 
	Y_t &=  B_t [  \lambda (Z_t - \EE[Z_t])  - \lambda^2 B^2/8  ] \sk
	\tilde{G}_n &=  \exp \Lp \sum_{t=n_0}^n Y_t \Rp \indic \{n \leq T \}.
	}
	So that $G$ can be written:
	\eqs{ 
	G_n = \tilde{G}_n \exp (- t_n ( \lambda \delta - \lambda^2 B^2/8 ) ).
	}
	Setting $\lambda = 4 \delta/B^2$:
	\eqs{ 
	G_n  = \tilde{G}_n \exp(-2 t_n \delta^2 / B^2).
	}
	Using the fact that $t_{\phi} \geq s$ if $\phi \leq T$,  we can upper bound $G_{\phi}$ by:
	\eqs{
	G_{\phi} =  \tilde{G}_{\phi} \exp(-2 t_{\phi} \delta^2 / B^2) \leq  \tilde{G}_{\phi}  \exp(-2 s \delta^2 / B^2).
	}
	It is noted that the above inequality holds even when $\phi = T+1$, since $G_{T+1} = \tilde{G}_{T+1} = 0$. Hence:
		\eqs{
	\EE[ G_{\phi} ] \leq \EE[ \tilde{G}_{\phi} ] \exp(-2 s \delta^2 / B^2).
	}
	We prove that $(\tilde{G}_n)_n$ is a super-martingale. We have that $\EE[ \tilde{G}_{T+1} | {\cal F}_T] = 0 \leq \tilde{G}_{T}$. For $n \leq T-1$, since $B_{n+1}$ is ${\cal F}_n$ measurable:
	\eqs{
	\EE[ \tilde{G}_{n+1} | {\cal F}_n] = \tilde{G}_n (  (1 - B_{n+1}) +   B_{n+1} \EE[ \exp(Y_{n+1}) ] )  . 
	}
	As proven by Hoeffding \cite{Hoeffding1963}[eq. 4.16] since $Z_{n+1} \in [0, B]$:
	\eqs{
	\EE[ \exp(\lambda (Z_{n+1} - \EE[ Z_{n+1}] ) ) ] \leq \exp(\lambda^2 B^2/8 ),
	}
 so $\EE[ \exp(Y_{n+1}) ] \leq 1$ and $(\tilde{G}_n)_n$ is indeed a supermartingale: $\EE[ \tilde{G}_{n+1} | {\cal F}_n] \leq \tilde{G}_n$. Since $\phi \leq T+1$ almost surely, and $(\tilde{G}_n)_n$ is a supermartingale, Doob's optional stopping theorem yields: $\EE[ \tilde{G}_{\phi}] \leq \EE[\tilde{G}_{n_0-1}] = 1$, and so
	\als{
		 \PP[& S_{\phi}  \geq t_{\phi} \delta  ,  \phi \leq T  ] \leq \mathbb{E}[G_\phi] \sk
		&  \le \EE[ \tilde{G}_{\phi} ] \exp(-2 s \delta^2 / B^2) \leq \exp(-2 s \delta^2 / B^2).
	}
	which concludes the proof. The second inequality is obtained by symmetry.

\ep

\subsection{Preliminary results}

Lemma~\ref{lem:deviation} states that if a set of instants $\Lambda$ can be decomposed into a family of subsets $(\Lambda(s))_{s\ge 1}$ of instants (each subset has at most one instant) where $k$ is tried sufficiently many times ($t_k(n) \geq \epsilon s$, for $n \in \Lambda(s)$), then the expected number of instants in $\Lambda$ at which the average reward of $k$ is badly estimated is finite.
 
\begin{lemma}\label{lem:deviation}
Let $k\in\{ 1,\ldots,K\}$, and $\epsilon > 0$. Define ${\cal F}_n$ the $\sigma$-algebra generated by $( X_k(t) )_{1 \leq t \leq n, 1 \leq k \leq K}$. Let $\Lambda \subset \NN$ be a (random) set of instants. Assume that there exists a sequence of (random) sets $(\Lambda(s))_{s\ge 1}$ such that (i) $\Lambda \subset \cup_{s \geq 1} \Lambda(s)$, (ii) for all $s\ge 1$ and all $n\in \Lambda(s)$, $t_k(n) \ge \epsilon s$, (iii) $|\Lambda(s)| \leq 1$, and (iv) the event $n \in \Lambda(s)$ is ${\cal F}_n$-measurable. Then for all $\delta > 0$:
\eq{\label{eq:ineq1}
\EE[ \sum_{n \geq 1} \indic\{ n \in \Lambda , | \hat\mu_k(n) - \EE[\hat\mu_k(n)] | > \delta \} ]  \leq  \frac{1}{\epsilon \delta^2}.
}
\end{lemma}
\bp
Let $T \geq 1$. For all $s \geq 1$, since $\Lambda(s)$ has at most one element, define $\phi_s = T+1$ if $\Lambda(s) \cap \{1 , \dots, T \}$ is empty and $\{ \phi_s \} = \Lambda(s)$ otherwise. Since $\Lambda \subset \cup_{s \geq 1} \Lambda(s)$, we have:
\als{
\sum_{n = 1}^{T} \indic\{ n \in \Lambda , | \hat\mu_k(n) - \EE[\hat\mu_k(n)]  | > \delta \} \sk 
\leq \sum_{s \geq 1} \indic \{ | \hat\mu_k(\phi_s) - \EE[\hat\mu_k(\phi_s)]  | > \delta , \phi_s \leq T \}.
}
Taking expectations:
\als{
\EE[ \sum_{n = 1}^{T} \indic\{ n \in \Lambda , | \hat\mu_k(n) - \EE[\hat\mu_k(n)]  | > \delta \} ] \sk 
\leq \sum_{s \geq 1} \PP [ | \hat\mu_k(\phi_s) - \EE[\hat\mu_k(\phi_s)]  | > \delta , \phi_s \leq T  ].
} 
 Since $\phi_s$ is a stopping time upper bounded by $T+1$, and that $t_k(\phi_s) \geq \epsilon s$ we can apply Lemma~\ref{lem:concentr} to obtain:
 \als{
\EE[ & \sum_{n = 1}^{T} \indic\{ n \in \Lambda , | \hat\mu_k(n) - \EE[\hat\mu_k(n)]  | > \delta \} ] \sk
&\leq  \sum_{s \geq 1} 2 \exp \Lp - 2 s \epsilon \delta^2 \Rp  \leq \frac{1}{\epsilon \delta^2}.
} 
  We have used the inequality: $\sum_{s \geq 1} e^{- s w} \leq \int_{0}^{+\infty} e^{- u w} du = 1/w$. Since the above reasoning is valid for all $T$, we obtain the claim~\eqref{eq:ineq1}.
\ep

A useful corollary of Lemma~\ref{lem:deviation} is obtained by choosing $\delta = \Delta_{k,k^\prime}/2$, when arms $k$ and $k^\prime$ are separated by at least $\Delta_{k,k^\prime}$.

\begin{lemma}\label{cor:deviation}
Let $k, k'\in\{ 1,\ldots,K\}$ with $k\neq k'$ and $\epsilon >0$. Define ${\cal F}_n$ the $\sigma$-algebra generated by $( X_k(t) )_{1 \leq t \leq n, 1 \leq k \leq K}$. Let $\Lambda \subset \NN$ be a (random) set of instants. Assume that there exists a sequence of (random) sets $(\Lambda(s))_{s\ge 1}$ such that (i) $\Lambda \subset \cup_{s \geq 1} \Lambda(s)$, (ii) for all $s\ge 1$ and all $n\in \Lambda(s)$, $t_k(n)\ge \epsilon s$ and $t_{k'}(n)\ge \epsilon s$, (iii) for all $s$ we have $|\Lambda(s)| \leq 1$ almost surely and (iv) for all $n \in \Lambda$, we have $\EE[\hat\mu_k(n)] \leq  \EE[\hat\mu_{k^\prime}(n)] - \Delta_{k,k^\prime}$ (v) the event $n \in \Lambda(s)$ is ${\cal F}_n$-measurable. Then:
\begin{equation}\label{eq:ineq2}
\EE[ \sum_{n \geq 1} \indic\{ n \in \Lambda , \hat\mu_k(n) > \hat\mu_{k^\prime}(n)  \} ] \leq \frac{ 8 }{\epsilon \Delta_{k,k^\prime}^2}.
\end{equation}
\end{lemma}

Lemma~\ref{lem:deviation_result} is straightforward from \cite{garivier2011}[Theorem 10]. It should be observed that this result is not a direct application of Sanov's theorem; Lemma \ref{lem:deviation_result} provides sharper bounds in certain cases, and it is also valid for non-Bernoulli distributed random variables.

\begin{lemma}\label{lem:deviation_result}
For $1 \leq t_k(n)  \leq \tau$ and $\delta > 0$, if $\{ X_k(i) \}_{1 \leq i \leq \tau}$ are independent random variables with mean $\mu_k$, we have that:
\als{
\PP \Lb t_k(n) I \Lp  \frac{1}{t_k(n)} \sum_{i=1}^{t_k(n)} X_k(i) , \mu_k \Rp \geq \delta  \Rb \sk 
\leq 2 e \ceil{\delta \log(\tau) }\exp(-\delta).	
}
\end{lemma}

We present results related to the KL divergence that will be instrumental when manipulating indexes $b_k(n)$. Lemma~\ref{lem:pinsker} gives an upper and a lower bound for the KL divergence. The lower bound is Pinsker's inequality. The upper bound is due to the fact that $I(p,q)$ is convex in its second argument.
\begin{lemma}\label{lem:pinsker}
For all $p,q \in [0,1]^2$, $p \leq q$:
\eq{\label{eq:pinkser_ineq}
 2 (p - q)^2 \leq I(p,q) \leq \frac{ (p-q)^2 }{q(1-q)}.
 }
 and 
 \eq{\label{eq:pinkser_equiv}
  I(p,q) \sim \frac{ (p-q)^2 }{q(1-q)} \;,\; q \to p^+
 }
 
\end{lemma}
\bp
	The lower bound is Pinsker's inequality. 
	For the upper bound, we have: 
	\eqs{
	\frac{\partial I}{\partial q}\Lp p,q \Rp = \frac{q-p}{q(1-q)}.
	}
	Since $q \mapsto \frac{\partial I}{\partial q}(p,q)$ is increasing, the fundamental theorem of calculus gives the announced result:
	\eqs{
	I(p,q) \leq \int_{p}^{q} \frac{\partial I}{\partial u} \Lp p,u \Rp du \leq \frac{(p-q)^2}{q(1-q)}. 
	}
	The equivalence comes from a Taylor development of $q \to I(p,q)$ at $p$, since: 
	\als{
	\frac{\partial I}{\partial q}(p,q) |_{q=p} = 0, \sk
	\frac{\partial^2 I}{\partial q^2}(p,q) |_{q=p} = \frac{1}{q(1-q)}.
	}
\ep	

We prove a deviation bound similar to that of Lemma~\ref{lem:deviation} for non-stationary environments.

\begin{lemma}\label{lem:deviation_ns}
Let $k\in\{ 1,\ldots,K\}$, $n_0 \in \NN$ and $\epsilon > 0$. Let $\Lambda \subset \NN$ be a (random) set of instants. Assume that there exists a sequence of (random) sets $(\Lambda(s))_{s\ge 1}$ such that (i) $\Lambda \subset \cup_{s \geq 1} \Lambda(s)$, (ii) for all $s\ge 1$ and all $n\in \Lambda(s)$, $t_k(n) \ge \epsilon s$, and (iii) for all $s \geq 1$ $|\Lambda(s) \cap [n_0,n_0+\tau] | \leq 1$. Then for all $\delta > 0$:
	\eqs{
\EE[ \sum_{n = n_0}^{n_0 + \tau} \indic\{ n \in \Lambda , | \hat\mu_k(n) - \EE[\hat\mu_k(n)] | > \delta \} ]  \leq  \frac{\log(\tau)}{2 \epsilon \delta^2} + 2.
}\end{lemma}

\bp Fix $s_0 \geq 1$. We use the following decomposition, depending on the value of $s$ with respect to $s_0$:
	\als{
	\{ n \in \Lambda , | \hat\mu_k(n) - \EE[\hat\mu_k(n)] | > \delta \} \subset A \cup B,
	}
where
\als{
A &= \{n_0,\dots,n_0 + \tau\} \cap ( \cup_{ 1 \leq s \leq s_0} \Lambda(s) ) , \sk
B &=  \{n_0,\dots,n_0 + \tau\}   \sk
  & \cap \{ n \in \cup_{ s \geq s_0} \Lambda(s) : | \hat\mu_k(n) - \EE[\hat\mu_k(n)] | > \delta  \}.
}
Since for all $s$, $|\Lambda(s) \cap \{n_0, \dots, n_0 + \tau \}|  \leq 1$, we have $|A| \leq  s_0$. The expected size of $B$ is upper bounded by:
	\als{
		E[ |B| ] &\leq \sum_{n = n_0}^{n_0 + \tau}  \PP[  n \in \cup_{s \geq s_0} \Lambda(s) , | \hat\mu_k(n) - \EE[\hat\mu_k(n)] | > \delta        ] \sk
						 &\leq \sum_{n = n_0}^{n_0 + \tau}  \PP[  | \hat\mu_k(n) - \EE[\hat\mu_k(n)] | > \delta   , t_k(n) \geq \epsilon s_0].
	}
For a given $n$, we apply Lemma~\ref{lem:concentr} with $n-\tau$ in place of $n_0$, and $\phi = n$ if $t_k(n) \geq \epsilon s_0$ and $\phi = T+1$ otherwise. It is noted that $\phi$ is indeed a stopping time. We get:
	\als{
	\PP[  | \hat\mu_k(n) - \EE[\hat\mu_k(n)] | > \delta   , t_k(n) \geq \epsilon s_0] \sk 
	\leq 2 \exp \Lp - 2 s_0 \epsilon \delta^2 \Rp.
	}
	Therefore, setting $s_0 =  \log(\tau) / (2 \epsilon \delta^2)$,
		\eqs{
		E[ |B| ] \leq 2 \tau \exp \Lp - 2 s_0 \epsilon \delta^2 \Rp = 2.
	}
Finally we obtain the announced result:
	\eqs{
\EE[ \sum_{n = n_0}^{n_0 + \tau} \indic\{ n \in \Lambda , | \hat\mu_k(n) - \EE[\hat\mu_k(n)] | > \delta \} ]  \leq   \frac{\log(\tau)}{2 \epsilon \delta^2} + 2.
}
	\ep
	
\begin{lemma}\label{cor:deviation_ns}
	Consider $k,k^\prime \in\{ 1,\ldots,K\}$, $n_0 \in \NN$ and $\epsilon > 0$. Let $\Lambda \subset \NN$ be a (random) set of instants. Assume that there exists a sequence of (random) sets $(\Lambda(s))_{s\ge 1}$ such that (i) $\Lambda \subset \cup_{s \geq 1} \Lambda(s)$, and (ii) for all $s\ge 1$ and all $n\in \Lambda(s)$, $t_k(n) \ge \epsilon s$, $t_{k^\prime}(n) \ge \epsilon s$ and (iii) for all $s \geq 1$ $|\Lambda(s) \cap [n_0,n_0+\tau] | \leq 1$ and (iv) for all $n \in \Lambda$, we have $\EE[\hat\mu_k(n)] \leq  \EE[\hat\mu_{k^\prime}(n)] - \Delta_{k,k^\prime}$.

Then for all $\delta > 0$:
	\eqs{
\EE[ \sum_{n = n_0}^{n_0 + \tau} \indic\{ n \in \Lambda , \hat\mu_k(n) > \hat\mu_{k^\prime}(n) \} ]  \leq  \frac{4 \log(\tau)}{ \epsilon \Delta_{k,k^\prime}^2} + 4.
}\end{lemma}

\section{Proofs for stationary environments}\label{sec:proof_stationary}

\subsection{Proof of Theorem~\ref{th:kluucb_finite} } 

{\bf Notations.} Throughout the proof, by a slight abuse of notation, we omit the floor/ceiling functions when it does not create ambiguity.	Consider a suboptimal arm $k \neq k^\star$. 
Define the difference between the average reward of $k$ and $k^\prime$ : $\Delta_{k,k^\prime} = | \mu_{k^\prime}  - \mu_{k} |  > 0$. We use the notation: 
\eqs{
t_{k,k^\prime}(n) = \sum_{t=1}^n \indic \{ L(t) = k, k(t) = k^\prime \}.
}
$t_{k,k^\prime}(n)$ is the number of times up time $n$ that $k^\prime$ has been selected given that $k$ was the leader.

\medskip
\noindent
{\bf Proof.} Let $T>0$. The regret $R^{\ouralgo}(T)$ of \ouralgo algorithm up to time $T$ is:
$$
R^{\ouralgo}(T) = \sum_{k\neq k^\star} (\mu_{k^\star} - \mu_k  ) \EE[ \sum_{n=1}^T   \indic\{ k(n)=k\} ].
$$
We use the following decomposition:
\als{
\indic\{ k(n)=k\}=\indic\{L(n)=k^\star,k(n)=k\} \sk
+\indic\{L(n)\neq k^\star,k(n)=k\}.
}
Now 
\begin{align*}
\sum_{k\neq k^\star}   & ( \mu_{k^\star} - \mu_k  ) \EE[ \sum_{n=1}^T   \indic\{ L(n)\neq k^\star, k(n)=k\} ]\\
& \le \sum_{k\neq k^\star}\EE[ \sum_{n=1}^T   \indic\{ L(n)\neq k^\star,k(n)=k\} ] \\
&\le \sum_{k\neq k^\star}\EE[ l_k(T)].
\end{align*}
Observing that when $L(n)=k^\star$, the algorithm selects a decision $(k,k^\star) \in E$, we deduce that: 
\begin{align*}
& R^{\ouralgo}(T) \leq  \sum_{k \neq k^\star } \EE[  l_k(T) ]   \sk 
& +  \sum_{(k,k^\star) \in E} ( \mu_{k^\star} - \mu_k  ) \EE[ \sum_{n=1}^T   \indic \{ L(n) = k^\star , k(n) = k \} ] 
\end{align*}

Then we analyze the two terms in the r.h.s. in the above inequality. The first term corresponds to the average number of times where $k^\star$ is not the leader, while the second term represents the accumulated regret when the leader is $k^\star$. The following result states that the first term is $O(\log(\log(T)))$:

\begin{theorem} \label{lemma:l1bound}
For $k \neq k^\star$, $ \EE[ l_k(T) ] = O(\log(\log(T)))$.
\end{theorem}

From the above theorem, we conclude that the leader is $k^\star$ except for a negligible number of instants (in expectation). When $k^\star$ is the leader, \ouralgo behaves as KL-UCB restricted to the set $N(k^\star)$ of possible decisions. Following the same analysis as in \cite{garivier2011} (the analysis of KL-UCB), we can show that for all $\epsilon > 0$ there are constants $C_1 \leq 7$ , $C_2(\epsilon)$ and $\beta(\epsilon) > 0$ such that:
\begin{align}
\EE[ & \sum_{n=1}^T  \indic \{L(n) = k^\star , k(n) = k \} ]  \sk 
&\leq  \EE[ \sum_{n=1}^T  \indic \{  b_{k}(n) \geq   b_{k^\star}(n) \} ]    \sk 
& \leq (1 + \epsilon) \frac{\log (T)}{I(\mu_k,\mu_{k^\star})}   + C_1 \log(\log(T)) + \frac{C_2(\epsilon)}{T^{\beta(\epsilon)}}.
\end{align}
Combining the above bound with Theorem \ref{lemma:l1bound}, we get:
\begin{align}\label{ref:regretlead2}
R^{\ouralgo}(T) & \leq   (1 + \epsilon) c(\theta) \log (T) +  O( \log(\log(T)) ),
\end{align}
which concludes the proof of Theorem~\ref{th:kluucb_finite}.
\ep

It remains to show that Theorem \ref{lemma:l1bound} holds, which is done in the next section. The proof of Theorem \ref{lemma:l1bound} is technical, and requires the concentration inequalities presented in section~\ref{sec:concentration_inequalities}. The theorem itself is proved in \ref{sec:pro}.

\subsection{Proof of Theorem~\ref{lemma:l1bound} }\label{sec:pro} 

Let $k$ be the index of a suboptimal arm. Let  $\delta > 0$, $\epsilon > 0$ small enough (we provide a more precise definition later on). We define $k_2 = \arg \max_{ k^\prime:(k,k^\prime) \in E } \mu_{k^\prime}$ the best neighbor of $k$. To derive an upper bound of $\mathbb{E}[l_k(T)]$, we decompose the set of times where $k$ is the leader into the following sets:
$$
\{n \leq T: L(n)=k\}\subset A_\epsilon\cup B_{\epsilon}^T,
$$
where
\begin{align*}
A_{\epsilon} & =  \{ n:	L(n) = k ,  t_{k_2}(n) \geq \epsilon l_k(n) \}\\
B_{\epsilon}^T &= \{  n \leq T: L(n) = k , t_{k_2}(n) \leq \epsilon l_k(n)\}.
\end{align*}
Hence we have:
\begin{equation*}
 \mathbb{E}[l_k(T)] \leq \mathbb{E}\big[ |A_{\epsilon}| + |B_{\epsilon}^T| \big],
\end{equation*}	
Next we provide upper bounds of $\mathbb{E}[ |A_{\epsilon}|]$ and $\mathbb{E}[ |B_{\epsilon}^T |]$. 

\medskip
\noindent
\underline{Bound on $\mathbb{E} |A_\epsilon|$.} Let $n \in A_{\epsilon}$ and assume that $l_k(n)=s$. By design of the algorithm, $t_{k}(n) \geq s/(\gamma+1)$. Also  $t_{k_2}(n) \geq \epsilon l_k(n) = \epsilon s$. We apply Lemma~\ref{cor:deviation} with $\Lambda(s) = \{ n  \in A_{\epsilon} , l_k(n) = s  \}$, $\Lambda =\cup_{s\ge 1}\Lambda(s)$. Of course, for any $s$, $|\Lambda(s)| \leq 1$. We have: $A_\epsilon =\{n\in \Lambda:\hat{\mu}_k(n)\ge \hat{\mu}_{k_2}(n)\}$, since when $n\in A_\epsilon$, $k$ is the leader. Lemma~\ref{cor:deviation} can be applied with $k'=k_2$. We get: $\EE| A_{\epsilon}|  < \infty$.

\medskip
\noindent
\underline{Bound on $\mathbb{E} |B_{\epsilon}^T|$.} We introduce the following sets:
\begin{itemize}
\item $C_\delta$ is the set of instants at which the average reward of the leader $k$ is badly estimated:
\eqs{
C_{\delta} = \{  n:  L(n) = k ,  | \hat{\mu}_k(n) - \mu_k | > \delta  \}.
}
\item $D_{\delta} = \cup_{ k^\prime \in N(k) \setminus \{ k_2 \}} D_{\delta,k^\prime}$ where $D_{\delta,k^\prime}=\{ n: L(n) = k , k(n) = k^\prime ,  | \hat{\mu}_{k^\prime}(n) - \mu_{k^\prime} | > \delta \}$  is the set of instants at which $k$ is the leader, $k^\prime$ is selected and the average reward of $k^\prime$ is badly estimated.
\item $E^T = \{ n \leq T: L(n) = k , b_{k_2}(n) \leq \mu_{k_2} \}$, is the set of instants at which $k$ is the leader, and the upper confidence index $b_{k_2}(n)$ underestimates the average reward $\mu_{k_2}$.
\end{itemize}
We first prove that $|B_{\epsilon}^T| \leq 2 \gamma (1 + \gamma) (|C_\delta| + |D_{\delta}| + |E^T|) + O(1)$ as $T$ grows large, and then provide upper bounds on $\mathbb{E}|C_\delta|$, $\mathbb{E}|D_\delta|$, and $\mathbb{E}|E^T|$. 
Let $n \in B_{\epsilon}^T$. When $k$ is the leader, the selected decision is in $N(k)$:
$$
l_k(n) = t_{k,k_2}(n) + \sum_{k' \in N(k) \setminus \{ k_2 \}} t_{k,k'}(n).
$$
We recall that $t_{k,k'}(n)$ denotes the number of times up to time $n$ when $k$ is the leader and $k'$ is selected. Since $n \in B_{\epsilon}^T$, $t_{k,k_2}(n)\le \epsilon l_k(n)$, from which we deduce that:
$$
(1 - \epsilon) l_k(n) \le  \sum_{k' \in N(k) \setminus \{ k_2 \}} t_{k,k'}(n) .
$$ 
Choose $\epsilon < 1/( 2( \gamma + 1) )$. With this choice, from the previous inequality, we must have that either (a) there exists $k_1 \in N(k) \setminus \{ k, k_2 \}$, $t_{k,k_1}(n) \geq l_k(n)/(\gamma + 1)$ or (b) $t_{k,k}(n) \geq (3/2) l_k(n)/( \gamma + 1) + 1$.

\medskip
(a) Assume that $t_{k,k_1}(n) \geq l_k(n)/( \gamma + 1)$.  Since $t_{k,k_1}(n)$ is only incremented when $k_1$ is selected and $k$ is the leader, and since $n \mapsto l_k(n)$ is increasing, there exists a unique $\phi(n) < n$ such that $L(\phi(n)) = k$, $k(\phi(n)) = k_1$, $t_{k,k_1}(\phi(n)) = \floor{ l_k(n)/(2( \gamma + 1))}$. $\phi(n)$ is indeed unique because $t_{k,k_1}(\phi(n))$ is incremented at time $\phi(n)$.
	
Next we prove by contradiction that for $l_k(n) \geq l_0$ large enough and $\delta$ small enough, we must have $\phi(n) \in C_\delta \cup D_{\delta} \cup E^T$. Assume that $\phi(n) \notin C_\delta \cup D_{\delta} \cup E^T$. Then $b_{k_2}(\phi(n)) \geq \mu_{k_2} $, $\hat\mu_{k_1}(\phi(n)) \leq \mu_{k_1} + \delta$. Using Pinsker's inequality and the fact that $t_{k_1}(\phi(n)) \geq t_{k,k_1}(\phi(n))$:
\begin{align*}
b_{k_1}(\phi(n)) & \leq \hat\mu_{k_1}(\phi(n)) \sk
&+ \sqrt{ \frac{ \log(l_k(\phi(n))) + c \log(\log(l_k(\phi(n))))   }{ 2 t_{k_1}(\phi(n)) }   } \sk
& \leq   \mu_{k_1} + \delta + \sqrt{\frac{ \log(l_k(n)) + c \log(\log(l_k(n)))}{ 2 \floor{ l_k(n)/(2( \gamma + 1))}}}.
\end{align*} 
Now select $\delta < (\mu_{k_2} - \mu_{k})/2$ and $l_0$ such that  $\sqrt{ (\log(l_0) + c \log(\log(l_0)))/ 2 \floor{ l_0/(2( \gamma + 1))}  }  \leq \delta$.
If $l_k(n) \geq l_0$: 
$$
b_{k_1}(\phi(n))\leq   \mu_{k_1}  + 2 \delta < \mu_{k_2} \leq b_{k_2}(\phi(n)),
$$
which implies that $k_1$ cannot be selected at time $\phi(n)$ (because $b_{k_1}(\phi(n)) < b_{k_2}(\phi(n))$), a contradiction.

\medskip	
(b) Assume that $t_{k,k}(n) \geq (3/2) l_k(n)/( \gamma + 1) + 1 = l_k(n)/( \gamma + 1) + l_k(n)/(2( \gamma + 1)) + 1$. There are at least $l_k(n)/(2( \gamma + 1)) + 1$ instants $\tilde{n}$ such that $l_{k}(\tilde{n}) - 1$ is not a multiple of $1/( \gamma + 1)$, $L(\tilde{n}) = k$ and $k(\tilde{n}) = k$. By the same reasoning as in (a) there exists a unique $\phi(n) < n$ such that $L(\phi(n)) = k$, $k(\phi(n)) = k$ , $t_{k,k}(\phi(n)) = \floor{l_k(n)/(2( \gamma + 1))}$ and $(l_k(\phi(n)) -1)$ is not a multiple of $1/( \gamma + 1)$. So $b_k(\phi(n)) \geq b_{k_2}(\phi(n))$. The same reasoning as that applied in (a) (replacing $k_1$ by $k$) yields $\phi(n) \in C_\delta \cup D_{\delta} \cup E^T$.

\medskip
 We define $B_{\epsilon,l_0}^T = \{ n: n \in B_{\epsilon}^T , l_k(n) \geq l_0 \}$, and we have that $|B_{\epsilon}^T| \leq l_0 +  |B_{\epsilon,l_0}^T|$. We have defined a mapping $\phi$ from $B_{\epsilon,l_0}^T$ to $C_{\delta} \cup D_{\delta} \cup E^T$. To bound the size of $B_{\epsilon,l_0}^T$, we use the following decomposition:
\als{
\{ & n: n \in B_{\epsilon,l_0}^T , l_k(n) \geq l_0 \} \sk
 & \subset \cup_{n^\prime \in  C_{\delta} \cup D_{\delta} \cup E^T} \{ n: n \in B_{\epsilon,l_0}^T , \phi(n) = n^\prime\}.
}
Let us fix $n^\prime$. If $n \in B_{\epsilon,l_0}^T$ and $\phi(n) = n^\prime$, then $\floor{l_k(n)/(2( \gamma + 1))} \in \cup_{k^\prime \in N(k) \setminus \{ k_2 \} }  \{ t_{k,k^\prime}(n^\prime) \}$ and $l_k(n)$ is incremented at time $n$ because $L(n) = k$. Therefore:
\eqs{
 | \{ n: n \in B_{\epsilon,l_0}^T , \phi(n) = n^\prime\} | \leq 2 \gamma( \gamma + 1).
 }
Using union bound, we obtain the desired result:
\eqs{
|B_{\epsilon}^T| \leq  l_0 +  |B_{\epsilon,l_0}^T| \leq   O(1) + 2 \gamma( \gamma + 1) ( |C_{\delta}| +  |D_{\delta}| + |E^T|).
}

\medskip
\noindent	
\underline{Bound on $\mathbb{E} |C_\delta|$.} We apply Lemma~\ref{lem:deviation} with $\Lambda(s) = \{ n: L(n)=k, l_k(n) = s \}$, and $\Lambda = \cup_{s\ge 1}\Lambda(s)$. Then of course, $|\Lambda(s)| \leq 1$ for all $s$. Moreover by design, $t_{k}(n) \geq s/(\gamma + 1)$ when $n \in \Lambda(s)$, so we can choose any $\epsilon < 1/( \gamma + 1)$ in Lemma~\ref{lem:deviation}. Now $C_{\delta}=\{n\in \Lambda: | \hat{\mu}_k(n) - \mu_k  | > \delta  \}$. From (\ref{eq:ineq1}), we get $\EE|C_{\delta}| < \infty$.

\medskip
\noindent
\underline{Bound on $\mathbb{E} |D_{\delta} |$.}
Let $k^\prime\in N(k) \setminus \{ k_2 \}$. Define for any $s$, $\Lambda(s) = \{ n : L(n)=k, k(n) = k^\prime , t_{k^\prime}(n) = s \}$, and $\Lambda=\cup_{s\ge 1}\Lambda(s)$. We have $|\Lambda(s)| \leq 1$, and for any $n\in \Lambda(s)$, $t_{k^\prime}(n)=s \ge \epsilon s$ for any $\epsilon <1$. We can now apply Lemma~\ref{lem:deviation} (where $k$ is replaced by $k^\prime$). Note that $D_{\delta,k^\prime} = \{n\in \Lambda: | \hat{\mu}_{k^\prime}(n) - \mu_{k^\prime}  | > \delta\}$, and hence (\ref{eq:ineq1}) leads to $\EE| D_{\delta,k^\prime}| <\infty$, and thus $\EE| D_{\delta}| <\infty$.

\medskip
\noindent
\underline{Bound on $\mathbb{E} |E^T |$.} 
We can show as in~\cite{garivier2011} (the analysis of KL-UCB) that $\mathbb{E}|E^T| = O(\log(\log(T)))$ (more precisely, this result is a simple application of Theorem 10 in ~\cite{garivier2011}). 

We have shown that $\mathbb{E} |B_{\epsilon}^T| = O(\log(\log(T)))$, and hence $\mathbb{E}[l_k(T)]=O(\log(\log(T)))$, which concludes the proof of Theorem \ref{lemma:l1bound}.
\ep

\section{Proofs for non-stationary environments}\label{sec:proof_non_stationary}

To simplify the notation, we remove the superscript $^{\tau}$ throughout the proofs, e.g $t_k^\tau(n)$ and $l_k^\tau(n)$ are denoted by $t_k(n)$ and $l_k(n)$.
\subsection{A lemma for sums over a sliding window} 
We will use Lemma~\ref{lem:window_sum} repeatedly to bound the number of times some events occur over a sliding window of size $\tau$. 
\begin{lemma}\label{lem:window_sum}
	Let $A \subset \NN$, and $\tau \in \NN$ fixed. Define $a(n) = \sum_{t=n-\tau}^{n-1} \indic \{ t \in A \}$. Then for all $T \in \NN$ and $s \in \NN$ we have the inequality:
	\eq{ \label{eq:wind}
	\sum_{n=1}^{T} \indic \{ n \in A ,  a(n) \leq s  \} \leq s \ceil{ T/\tau }.
	}

As a consequence, for all $k \in \{1 , \dots , K \}$, we have:
	\al{
	\sum_{n=1}^{T} \indic \{ k(n) = k ,  t_k(n) \leq s  \} & \leq s \ceil{ T/\tau }, \label{eq:window_sum1} \\
	\sum_{n=1}^{T} \indic \{ L(n) = k ,  l_k(n) \leq s  \} & \leq s \ceil{ T/\tau }. \nonumber
	}
These inequalities are obtained by choosing $A = \{ n : k(n) = k \}$ and $A = \{ n : L(n) = k \}$ in (\ref{eq:wind}).
\end{lemma}
\bp
We decompose $\{1 , \dots , T \}$ into intervals of size $\tau$: $\{1 , \dots , \tau \}$ , $\{\tau + 1 , \dots , 2 \tau \}$ etc. We have:
	\al{
	\sum_{n=1}^{T} & \indic\{   n \in A ,  a(n) \leq s  \} \sk 
	&\leq \sum_{i=0}^{ \ceil{ T/\tau }-1 } \sum_{n=1}^{\tau} \indic\{ n + i \tau \in A , a(n + i \tau) \leq s   \label{eq:time_windows}   \}. 
	}
Fix $i$ and assume that $\sum_{n=1}^{\tau} \indic\{ n + i \tau \in A , a(n + i \tau) \leq s  \} > s$. Then there must exist $n^\prime < \tau$ such that $n^\prime \in A$ and $\sum_{n=1}^{n^\prime} \indic\{ n + i \tau \in A , a(n + i \tau) \leq s  \}  = s$. Since $a(n^\prime + i \tau) \geq \sum_{n=1}^{n^\prime} \indic\{ n + i \tau \in A , a(n + i \tau) \leq s  \}$, we have $a(n^\prime + i \tau) \geq s$. As $n^\prime \in A$, we must have $a(n^{\prime\prime} + i \tau) \geq (s+1)$ for all $n^{\prime\prime} > n^\prime$ such that $n^{\prime\prime} \in A$. So 
	\als{
	\sum_{n=1}^{\tau} \indic\{ n + i \tau  \in A , a(n + i \tau) \leq s  \} \sk
	=  \sum_{n=1}^{n^\prime} \indic\{ n + i \tau  \in A , a(n + i \tau) \leq s  \} = s,
	}
	which is a contradiction. Hence, for all $i$:
	\eqs{
	\sum_{n=1}^{\tau} \indic\{ n + i \tau  \in A , a(n + i \tau) \leq s  \} \leq s,
	}
and substituting in~\eqref{eq:time_windows} gives the desired result:
\eqs{
	\sum_{n=1}^{T} \indic\{  n \in A , a(n) \leq s  \} \leq \sum_{i=0}^{\ceil{ T/\tau }-1 } s =  s \ceil{ T/\tau }.
}
\ep

\subsection{Regret of SW-KL-UCB}

	In order to analyze the regret of \ouralgosw, we first have to analyze the regret SW-KL-UCB on which \ouralgosw is based.
\begin{theorem}	\label{th:regretSWKLUCB}
Let $\Delta$: $2\tau\sigma < \Delta < \Delta_0$. Assume that for any $n\ge 1$, $\mu^\star(n)\in [a,1-a]$ for some $a>0$. Further suppose that $\mu_k(\cdot)$ is $\sigma$-Lipschitz for any $k$. The regret per unit time under $\pi=$SW-KL-UCB with a sliding window of size $\tau$ satisfies: if $a>\sigma\tau$, then for any $T\ge 1$,
\als{	
\frac{R^{\pi}(T)}{T} &\leq  \frac{H(\Delta,T)}{T} \Delta \sk
 &+ K \Lp 1 + g_0^{-1/2} \Rp \frac{\log(\tau) + c \log(\log(\tau)) + C_1}{2 \tau (\Delta - 2 \tau \sigma)^2} ,
}
where $C_1$ is a positive constant and $g_0 = (a - \sigma\tau)(1-a + \sigma\tau)/2$.
\end{theorem}

 Recall that due to the changing environment and the use of a sliding window, the empirical reward is a biased estimator of the average reward, and that its bias is upper bounded by $\sigma \tau$. 

To ease the regret analysis, we first provide bounds on the empirical reward. Unlike in the stationary case, the empirical reward $\hat\mu_{k}(n)$ is not a sum of $t_k(n)$ i.i.d. variables. We define $\overline{X}_k(n^\prime,n) = X_k(n^\prime) + ( \mu_{k}(n) + \sigma |n^\prime - n | - \mu_{k}(n^\prime)     )$ , $\underline{X}_k(n^\prime,n) = X_k(n^\prime) + ( \mu_{k}(n) - \sigma |n^\prime - n | - \mu_{k}(n^\prime))$ and:
\als{
\underline{\hat\mu}_{k}(n)  &= \frac{1}{t_k(n)} \sum_{ n^\prime = n-\tau }^{n}  \underline{X}_k(n^\prime,n) \indic \{k(n^\prime) = k \},\sk
\overline{\hat\mu}_{k}(n)  &= \frac{1}{t_k(n)} \sum_{ n^\prime = n-\tau }^{n}   \overline{X}_k(n^\prime,n) \indic \{k(n^\prime) = k \}.
}
Then of course, $\underline{\hat\mu}_{k}(n) \leq {\hat\mu}_{k}(n) \leq \overline{\hat\mu}_{k}(n)$.
	
\medskip	
Now the regret under $\pi$=\ouralgosw is given by:
\eqs{
R^{\pi}(T) = \sum_{n=1}^T \sum_{k=1}^K (  \mu_{k^\star}(n) - \mu_k(n)  ) \PP [ k(n) = k ]. 
}
We define $I_{min} = 2(\Delta - 2\tau\sigma)^2$. Let $\epsilon >0$ and ${\cal K}^\tau = (1+\epsilon) \frac{  \log(\tau) + c \log(\log(\tau))}{I_{\min }}$. We introduce the following sets of events: 

\medskip
\noindent
(i) $A  = \cup_{k=1}^{K} A_k $, where
\als{
A_{k} = \{ & 1 \leq n \leq T : k(n) = k , | \mu_k(n) -  \mu_{k^\star}(n) |  <  \Delta \},
}
$A_k$ is the set of times at which $k$ is chosen, and $k$ is "close" to the optimal decision. Note that, by definition, $|A| \leq H(\Delta,T)$.

\medskip
\noindent
(ii)  $B = \{ 1 \leq n \leq T : b_{k^\star}(n) \leq   \mu_{k^\star}(n) - \tau \sigma  \}$. $B$ is the set of times at which the index $b_{k^\star}(n)$ underestimates the average reward of the optimal decision (with an error greater than the bias $\tau \sigma$).

\medskip
\noindent
(iii) $C =  \cup_{k=1}^{K} C_k$ ,  $C_k = \{ 1 \leq n \leq T : k(n) = k , t_k(n) \leq {\cal K}^\tau \}$. $C_k$ is the set of times at which $k$ is selected and it has been tried less than ${\cal K}^\tau$ times.

\medskip
\noindent
(iv) $D =  \cup_{k=1}^{K} D_k$, $D_k = \{ 1 \leq n \leq T : k(n) = k, n \notin (A \cup B \cup C) \}$. $D_k$ is the set of times where (a) $k$ is chosen, (b) $k$ has been tried more than ${\cal K}^\tau$ times, (c) $k$ is not close to the optimal decision, and (d) the average reward of the optimal decision is not underestimated.

\medskip
We will show that:
\begin{equation}\label{eq:i}
  \sum_{n \in A} (\mu^*(n) - \mu_{k(n)}(n)) \leq \Delta H(\Delta,T). 
\end{equation}
and the following inequalities
\als{
\EE[|B|] & \leq O(T/\tau), \quad \EE[|C_k|] \leq {\cal K}^\tau \ceil{T/\tau}, \sk
\EE[|D_k]] & \leq  \frac{T}{(\tau \log(\tau)^c)^{g_0 \epsilon^2}} . }
We deduce that:
\als{
R^{\pi}(T) & \leq  \Delta H(\Delta,T)  + O(T/\tau)  \sk
&+ K {\cal K}^\tau \floor{T/\tau} + \frac{K T}{(\tau \log(\tau)^c)^{g_0 \epsilon^2}} ,
}
which proves Theorem~\ref{th:regretSWKLUCB}.

\medskip
\noindent
\underline{Proof of (\ref{eq:i}).} Let $n \in A_k$. If $n \in A_{k}$, by definition we have $|\mu_{k^\star}(n) - \mu_k(n) |  <  \Delta$. Then if $k(n) = k$, we have that $\mu^*(n) - \mu_{k(n)}(n) \leq \Delta$ so that:
\eqs{ 
  \sum_{n \in A} (\mu^*(n) - \mu_{k(n)}(n)) \leq  \Delta|A| \leq \Delta H(\Delta,T),
}
which completes the proof of (\ref{eq:i}).

\medskip
\noindent
\underline{Bound on $\mathbb{E}[|B|]$.} Let $n\in B$. Note that $\underline{\hat\mu}_{k^\star}(n) \leq {\hat\mu}_{k^\star}(n)\le b_{k^\star}(n)$. Since $b_{k^\star}(n) \leq   \mu_{k^\star}(n) - \sigma \tau$, we deduce that: $\underline{\hat\mu}_{k^\star}(n) \leq  \mu_{k^\star}(n) - \sigma \tau$. Now we have:
\als{
\PP [ n \in B] &=  \PP [ b_{k^\star}(n) \leq   \mu_{k^\star}(n) - \sigma \tau  ] \sk
& =  \PP[ t_{k^\star}(n) I \Lp \hat\mu_{k^\star}(n) , \mu_{k^\star}(n) - \sigma \tau \Rp \sk
& \hspace{2cm} \geq \log(\tau) + c \log(\log(\tau)) ] \sk
& \stackrel{(a)}{\leq}  \PP [ t_{k^\star}(n) I \Lp \underline{\hat\mu}_{k^\star}(n) , \mu_{k^\star}(n) - \sigma \tau \Rp \sk
& \hspace{2cm} \geq \log(\tau) + c \log(\log(\tau)) ] \sk
&\stackrel{(b)}{\leq} \frac{2 e}{\tau (\log(\tau))^{c-2}},  
}
where (a) is due to the fact that $\underline{\hat\mu}_{k^\star}(n) \leq {\hat\mu}_{k^\star}(n)$, and (b) is obtained applying Lemma \ref{lem:deviation_result}. Hence: $\EE [|B|] \leq O(T/\tau)$.
	
\medskip	
\noindent
\underline{Bound on $\mathbb{E}[|C_k|]$.} Using Lemma~\ref{lem:window_sum}, we get $|C_k| \leq   {\cal K}^\tau \ceil{ T/\tau }$, and hence $|C| \leq  K {\cal K}^\tau \floor{ T/\tau }$.

\medskip
\noindent
\underline{Bound on $\mathbb{E}[|D_k|]$.} We will prove that $n \in D_k$ implies that $\overline{\hat\mu}_{k}(n)$ deviates from its expectation by at least $f(\epsilon,I_{min}) > 0$ so that:
\eqs{ \PP[ n \in D_k ] \leq \PP \Lb \overline{\hat\mu}_{k}(n) - \EE[ \overline{\hat\mu}_{k}(n)] > f(\epsilon,I_{\min}) \Rb. }

Let $n \in D_k$. Since $k(n) = k$ and $b_{k^\star}(n) \geq \mu_{k^\star}(n) - \sigma \tau$, we have $b_k(n) \geq \mu_{k^\star}(n) - \sigma \tau$. We decompose $D_k$ as follows:
\als{
	D_{k} &= D_{k,1} \cup D_{k,2} \sk
	D_{k,1} &= \{ n \in D_k:  \overline{\hat\mu}_{k}(n) \geq   \mu_{k^\star}(n) - \sigma \tau   \} \sk
	D_{k,2} &= \{ n \in D_k:  \overline{\hat\mu}_{k}(n) \leq    \mu_{k^\star}(n) - \sigma \tau   \}
} 
	If $n \in D_{k,1}$, $\overline{\hat\mu}_{k}(n) - \EE[ \overline{\hat\mu}_{k}(n)] \geq  \mu_{k^\star}(n) -\mu_{k}(n) - 2\sigma \tau > 0$ so that $\overline{\hat\mu}_{k}(n)$ indeed deviates from its expectation. Now let $n \in D_{k,2}$. We have:
\als{
& \PP[ n \in D_{k,2} ] \sk
& \leq  \PP[ b_k(n) \geq  \mu_{k^\star}(n) - \sigma \tau , n \in D_{k,2}] \sk
& =  \PP[ t_k(n)I \Lp {\hat\mu}_{k}(n)   ,   \mu_{k^\star}(n) - \sigma \tau   \Rp \sk
& \hspace{2cm} \leq  \log(\tau) + c \log(\log(\tau))      , n \in D_{k,2}] \sk
& \stackrel{(a)}{\leq} \PP [  {\cal K}^\tau I \Lp  \overline{\hat\mu}_{k}(n) ,   \mu_{k^\star}(n) - \sigma \tau   \Rp \sk
& \hspace{2cm} \leq  \log(\tau) + c \log(\log(\tau)),  t_k(n) \geq {\cal K}^\tau ] \sk
&= \PP \Lb I \Lp \overline{\hat\mu}_{k}(n) ,  \mu_{k^\star}(n) - \sigma \tau  \Rp \leq \frac{I_{\min}}{1+\epsilon} , t_k(n) \geq {\cal K}^\tau\Rb ,
}
where in (a), we used the facts that: $\overline{\hat\mu}_{k}(n) \leq    \mu_{k^\star}(n) - \sigma \tau$, $\overline{\hat\mu}_{k}(n) \geq {\hat\mu}_{k}(n)$, and $t_k(n)\ge {\cal K}^\tau$ ($n\notin C$). It is noted that since $n \notin A_k$, by Pinkser's inequality we have that: $I( \mu_k(n) + \tau\sigma , \mu_{k^\star}(n) - \tau\sigma) \geq 2 (\mu_{k^\star}(n) - \mu_k(n) - 2\tau\sigma )^2  \geq  2 (\Delta - 2 \tau\sigma )^2 = I_{min}$. By continuity and monotonicity of the KL divergence, there exists a unique positive function $f$ such that: 
	\als{
	I \Lp \mu_{k}(n) + \sigma \tau + f(\epsilon,I_{\min} )  ,  \mu_{k^\star}(n) - \sigma \tau  \Rp = \frac{I_{min}}{1+\epsilon},\sk
	\mu_{k}(n) + \sigma \tau + f(\epsilon,I_{\min} ) \leq  \mu_{k^\star}(n) - \sigma \tau.
	}
	
	We are interested in the asymptotic behavior of $f$ when $\epsilon$ , $I_{min}$ both tend to $0$ .  Define $\mu^\prime$ , $\mu^{\prime\prime}$ and $\mu_0$ such that
	\eqs{  \mu_{k}(n) + \sigma \tau  \leq \mu^\prime  \leq \mu^{\prime\prime} \leq \mu_0 =  \mu_{k^\star}(n) - \sigma \tau.}
	and
	\eqs{
	I( \mu^\prime, \mu_0) = I_{\min} \;\;, \;\;
	I( \mu^{\prime\prime}, \mu_0) = \frac{I_{\min}}{1 + \epsilon}.
	}
	Using the equivalent \eqref{eq:pinkser_equiv} given in Lemma~\ref{lem:pinsker}, there exists a function $a$ such that:
	\als{
		\frac{ ( \mu_0 - \mu^\prime )^2  }{ \mu_0(1-\mu_0) }(1 + a( \mu_0 - \mu^\prime))  &= I_{\min}, \sk
		\frac{ ( \mu_0 - \mu^{\prime\prime} )^2  }{ \mu_0(1-\mu_0) }(1 + a( \mu_0 - \mu^{\prime\prime} ) )  &= \frac{I_{\min}}{1 + \epsilon}.
	}
 with $a(\delta) \to 0$ when $\delta \to 0^+$. It is noted that $0 \leq \mu_0 - \mu^{\prime\prime} \leq \mu_0 - \mu^{\prime} = o(1)$ when $I_{\min} \to 0^{+}$ by continuity of the KL divergence. Hence:
	\eqs{
	 \mu^{\prime\prime} - \mu^\prime =  \Lp \frac{\epsilon}{2} + o(1) \Rp \sqrt{ \mu_0(1-\mu_0) I_{\min} }.
	}
	Using the inequality
	\als{
	f(\epsilon,I_{min} ) &=  \mu^{\prime\prime} - (\mu_{k}(n) + \sigma \tau)  \sk 
	&\geq \mu^{\prime\prime} - \mu^{\prime} = \frac{\epsilon}{2} \sqrt{ \mu_0(1-\mu_0) I_{\min} },
	}
	we have proved that:	
	\als{
		2 f(\epsilon,I_{min} )^2 \geq \epsilon^2 g_0 I_{\min} + o(\epsilon^2)
		}
	with
	\eqs{
	g_0 = (a - \sigma \tau )(1 -  a + \sigma \tau)/2. 
	}
	
	Therefore, since $\EE[ \overline{\hat\mu}_{k}(n)] \leq \mu_{k}(n) + \sigma \tau$, as claimed, we have
	\als{
	& \PP[ n \in D_{k}] \sk 
	& \leq \PP \Lb \overline{\hat\mu}_{k}(n) - \EE[ \overline{\hat\mu}_{k}(n)]  \geq f(\epsilon,I_{\min}) \;,\; t_k(n) \geq {\cal K}^\tau \Rb.
	}

	We now apply Lemma~\ref{lem:concentr} with $n-\tau$ in place of $n_0$, ${\cal K}^\tau$ in place of $s$ and $\phi = n$ if $t_k(n) \geq {\cal K}^\tau$ and $\phi = T+1$ otherwise. We obtain, for all $n$: 
\als{
	 \PP[& n \in D_k]  \sk
	 &\leq \PP \Lb \overline{\hat\mu}_{k}(n) - \EE[ \overline{\hat\mu}_{k}(n)]  \geq f(\epsilon,I_{\min}) ,  t_k(n) \geq {\cal K}^\tau \Rb \sk
	&\leq \exp \Lp - 2 {\cal K}^\tau f(\epsilon,I_{\min})^2 \Rp \leq \frac{1}{(\tau \log(\tau)^c)^{g_0 \epsilon^2}},
	}	
	and we get the desired bound by summing over $n$:
	\eqs{
	\EE[|D_k|] = \sum_{n=1}^T \PP[ n \in D_k] \leq \frac{T}{(\tau \log(\tau)^c)^{g_0 \epsilon^2}}.
	}

\subsection{Proof of Theorem~\ref{th:nonstat} } 	

We first introduce some notations. For any set $A$ of instants, we use the notation: $A[n_0,n] = A \cap \{ n_0,\dots,n_0+\tau \}$. Let $n_0 \leq n$. We define $t_k(n_0,n)$ the number of times $k$ has been chosen during interval $\{ n_0,\dots,n_0+\tau \}$, $l_k(n_0,n)$ the number of times $k$ has been the leader, and $t_{k,k^\prime}(n_0,n)$ the number of times $k ^\prime$ has been chosen while $k$ was the leader:
\als{
t_k(n_0,n) &= \sum_{n^\prime=n_0}^n \indic \{ k(n^\prime) = k \},\sk
l_k(n_0,n) &= \sum_{n^\prime=n_0}^n \indic \{ L(n^\prime) = k \}, \sk
t_{k,k^\prime}(n_0,n) &= \sum_{n^\prime=n_0}^n \indic \{ L(n^\prime) = k, k(n^\prime) = k^\prime \}.
}
Note that $l_k(n-\tau,n) = l_k(n)$, $t_k(n-\tau,n) = t_k(n)$ and $t_{k,k^\prime}(n-\tau,n) = t_{k,k^\prime}(n)$. Given $\Delta > 0$, we define the set of instants at which the average reward of $k$ is separated from the average reward of its neighbours by at least $\Delta$:
\eqs{
	{\cal N}_k(\Delta) = \cap_{   (k^\prime,k) \in E } \{ n: | \mu_k(n) - \mu_{k^\prime}(n)  | > \Delta  \}.
}
We further define the amount of time that $k$ is suboptimal, $k$ is the leader, and it is well separated from its neighbors:
\eqs{
{\cal L}_k(\Delta) = \{ n: L(n) = k \neq k^\star(n), n \in {\cal N}_k(\Delta) \}.
}

By definition of the regret under $\pi=$\ouralgosw:
	\eqs{
		R^{\pi}(T) = \sum_{n=1}^T \sum_{k \neq k^{\star}(n)} (\mu_{k^\star}(n)  - \mu_k(n)) \PP [ k(n) = k ].
	}
To bound the regret, as in the stationary case, we split the regret into two components: the regret accumulated when the leader is the optimal arm, and the regret generated when the leader is not the optimal arm. The regret when the leader is suboptimal satisfies: 
\als{
\sum_{n=1}^T & \sum_{k \neq k^{\star}(n)} (\mu_{k^\star}(n)  - \mu_k) \indic\{ k(n) = k  , L(n) \neq k^{\star}(n)\} \sk 
&\leq \sum_{n=1}^T \indic\{ L(n) \neq k^{\star}(n)\} \sk
&\leq \sum_{n=1}^T \sum_{k \neq k^{\star}(n)} \indic\{ L(n)=k \neq k^{\star}(n)\} \sk
&\leq  \sum_{n=1}^T \sum_{k \neq k^{\star}(n)}  \indic\{ n \in {\cal L}_k(\Delta) \} \sk 
&+ \indic\{ \exists k^{\prime} : (k,k^\star) \in E: | \mu_{k}(n) -  \mu_{k^\prime}(n) | \leq \Delta  \} \sk
&\leq \Lp \sum_{k=1}^K |{\cal L}_k(\Delta)[0,T]| + H(\Delta,T)  \Rp.
}
		
Therefore the regret satisfies:
\al{\label{eq:reget_non_stat}
R^{\pi}(T) &\leq  \Lp H(\Delta,T) + \sum_{k=1}^K \EE[|{\cal L}_k(\Delta)[0,T]|] \Rp \sk
&+ \sum_{n=1}^T \sum_{(k,k^\star(n)) \in E}  (\mu_{k^\star}(n)  - \mu_{k}(n) ) \PP[ k(n) = k ].
}
The second term of the r.h.s in \eqref{eq:reget_non_stat} is the regret of \ouralgosw when $k^{\star}(n)$ is the leader. This term can be analyzed using the same techniques as those used for the analysis of SW-KL-UCB and is upper bounded by the regret of SW-KL-UCB. It remains to bound the first term of the r.h.s in \eqref{eq:reget_non_stat}.

\begin{theorem}\label{th:kluucb_chang_leader}
	Consider $\Delta > 4 \tau \sigma$. Then for all $k$: 
\begin{equation}\label{eq:reget_non_stat2}
\EE[|{\cal L}_k(\Delta)[0,T]|] \leq C_1\times \frac{T \log(\tau)}{\tau (\Delta - 4 \tau \sigma)^2},
\end{equation}
where $C_1>0$ does not depend on $T$, $\tau$, $\sigma$ and $\Delta$.
\end{theorem}

Substituting~\eqref{eq:reget_non_stat2} in~\eqref{eq:reget_non_stat}, we obtain the announced result.

\ep

\subsection{Proof of Theorem~\ref{th:kluucb_chang_leader}} 

It remains to prove Theorem~\ref{th:kluucb_chang_leader}. Define $\delta = (\Delta - 4 \tau \sigma)/2$. We can decompose $\{1,\dots,T\}$ into at most $\ceil{T/\tau}$ intervals of size $\tau$. Therefore, to prove the theorem, it is sufficient to prove that for all $n_0 \in {\cal L}_k(\Delta)$ we have:
\eqs{
\EE[|{\cal L}_k(\Delta)[n_0,n_0+\tau] |] \leq O \Lp \frac{\log(\tau)}{\delta^2} \Rp.
} 
	
In the remaining of the proof, we consider an interval $\{ n_0,\dots,n_0+\tau \}$, with $n_0 \in {\cal L}_k(\Delta)$ fixed. It is noted that the best neighbour of $k$ changes with time. We define $k_2(n)$ the best neighbor of $k$ at time $n$. From the Lipschitz assumption and the fact that $\Delta > 4 \tau \sigma$, we have that for all $n \in \{ n_0,\dots,n_0+\tau \}$, $k_2(n) = k_2(n_0)$. Indeed for all $n \in \{ n_0,\dots,n_0+\tau \}$:

\als{
	 &  \mu_{k_2(n_0)}(n)  - \mu_{k}(n)   \sk 
	 & \geq \mu_{k_2(n_0)}(n_0)  - \mu_{k}(n_0)  - 2 (n-n_0) \sigma \sk
	 & \geq \Delta - 2 \tau \sigma \geq  2 \tau \sigma > 0.
}
We write $k_2=k_2(n_0) = k_2(n)$ when this does not create ambiguity. We will use the fact that, for all $n \in \{ n_0,\dots,n_0+\tau \}$:
\als{
\EE[\hat\mu_{k_2}(n)] - \EE[\hat\mu_k(n)]  &\geq   \mu_{k_2}(n) -  \mu_{k}(n)  - 2 \tau \sigma, \sk
&\geq 	 \mu_{k_2}(n_0) - \mu_{k}(n_0)  - 4 \tau \sigma, \sk
&\geq \Delta  - 4 \tau \sigma  = 2 \delta > 0.
}

We decompose ${\cal L}_k(\Delta)[n_0,n_0+\tau] = A_{\epsilon}^{n_0} \cup B_{\epsilon}^{n_0}$, with:
\begin{itemize}
\item[] $A_{\epsilon}^{n_0} = \{  n \in {\cal L}_k(\Delta)[n_0,n_0+\tau] , t_{k_2}(n) \geq \epsilon l_k(n_0,n) \}$ the set of times where $k$ is the leader, $k$ is not the optimal arm, and its best neighbor $k_2$ has been tried sufficiently many times during interval $\{ n_0,\dots,n_0+\tau \}$,
\item[] $B_{\epsilon}^{n_0} = \{ n \in {\cal L}_k(\Delta)[n_0,n_0+\tau] , t_{k_2}(n) \leq \epsilon l_k(n_0,n) \}$ the set of times where $k$ is the leader, $k$ is not the optimal arm, and its best neighbor $k_2$ has been little tried during interval $\{ n_0,\dots,n_0+\tau \}$.
\end{itemize}
	
\medskip
\noindent
\underline{Bound on $\mathbb{E}[A_{\epsilon}^{n_0}]$.} Let $n \in A_{\epsilon}^{n_0}$. We recall that $\EE[\hat\mu_{k_2}(n)] - \EE[\hat\mu_k(n)]  \geq 2 \delta$, so that the reward of $k$ or $k_2$ must be badly estimated at time $n$:
	\als{
	\PP[ &n \in A_{\epsilon}^{n_0}] \sk 
	&\leq \PP[ |\hat\mu_{k}(n) - \EE[\hat\mu_{k}(n)]| > \delta] \sk
	&+ \PP[ |\hat\mu_{k_2}(n) - \EE[\hat\mu_{k_2}(n)]| > \delta].
	}
We apply Lemma~\ref{cor:deviation_ns}, with $k^\prime = k_2$,  $\Delta_{k,k^\prime} = 2 \delta$, $\Lambda(s) = \{ n \in A_{\epsilon}^{n_0}, l_k(n_0,n) =  s\}$, $t_{k_2}(n) \geq \epsilon l_k(n_0,n)  = \epsilon s$. By design of \ouralgosw : $t_k(n) \geq l_k(n_0,n)/(\gamma + 1) = s/(\gamma + 1)$. Using the fact that $|\Lambda(s)| \leq 1$ for all $s$, we have that:
\eqs{
\EE[ A_{\epsilon}^{n_0} ] \leq O \Lp \frac{\log(\tau)}{\epsilon \delta^2} \Rp.
}
	
\medskip
\noindent
\underline{Bound on $\mathbb{E}[B_{\epsilon}^{n_0}]$.} Define $l_0$ such that 
\eqs{
\sqrt{ \frac{ \log(l_0) + c \log(\log(l_0)) }{2 \floor{l_0/(2(\gamma + 1))}} } \leq \delta.
} 
In particular we can choose $l_0 = 2(\gamma + 1) (\log(1/\delta)/\delta^2) $. Indeed, with such a choice we have that 
\eqs{
\sqrt{ \frac{ \log(l_0) + c \log(\log(l_0)) }{2 \floor{l_0/(2(\gamma + 1))}}} \sim \delta/2 \;,\; \delta \to 0^+.
}
Let $\epsilon < 1/(2(\gamma + 1))$, and define the following sets:
\begin{itemize}
\item[] $C_\delta^{n_0}$ is the set of instants at which the average reward of the leader $k$ is badly estimated:
\als{
C_{\delta}^{n_0} &= \{ n \in \{ n_0,\dots,n_0+\tau \} \sk
&:  L(n) = k \neq k^\star(n),  |\hat{\mu}_k(n) - \EE[\hat{\mu}_k(n)]| > \delta  \};
}
\item[] $D_{\delta}^{n_0} = \cup_{k^\prime \in N(k) \setminus \{ k_2 \} }  D_{\delta,k^\prime}^{n_0} $ where $D_{\delta,k'}^{n_0}=\{ n: L(n) = k \neq k^\star(n), k(n) = k^\prime, | \hat{\mu}_{k^\prime}(n) - \EE[\hat{\mu}_{k^\prime}(n)] | > \delta \}$. $D_{\delta}^{n_0}$ is the set of instants at which $k$ is the leader, $k^\prime$ is selected and the average reward of $k^\prime$ is badly estimated.
\item[] $E^{n_0} = \{ n \leq T: L(n) = k \neq k^\star(n), b_{k_2}(n) \leq \EE[\hat{\mu}_{k_2}(n)] \}$ is the set of instants at which $k$ is the leader, and the upper confidence index $b_{k_2}(n)$ underestimates the average reward $\EE[\hat{\mu}_{k_2}(n)]$.
\end{itemize}
	
Let $n \in B_{\epsilon}^{n_0}$. Write $s = l_k(n_0,n)$, and we assume that $s \geq l_0$. Since $t_{k_2}(n_0,n) \leq \epsilon l_k(n_0,n)$ and the fact that $l_{k}(n_0,n) = t_{k_2}(n_0,n) + \sum_{k' \in N(k) \setminus \{ k_2 \} } t_{k'}(n_0,n) $, we must have (a) there exists $k_1 \in N(k) \setminus \{ k,k_2 \}$ such that $t_{k_1}(n_0,n) \geq s/(\gamma + 1)$ or  (b) $t_{k_1}(n_0,n) \geq (3/2)s/(\gamma + 1) + 1$. Since $t_{k,k}(n)$ and $t_{k,k_2}(n)$ are incremented only at times when $k(n) = k$ and $k(n) = k_2$ respectively, there must exist a unique index $\phi(n) \in \{ n_0,\dots,n_0+\tau \}$ such that either: (a) $t_{k,k_1}(\phi(n)) = \floor{s/(2(\gamma + 1))}$  and $k(\phi(n)) = k_1$; or (b) $t_{k,k_2}(\phi(n)) = \floor{(3/2)s/(\gamma + 1)}$ and  $k(n) = k$ and $l_k(\phi(n))$ is not a multiple of $3$. In both cases, as in the proof of theorem~\ref{lemma:l1bound}, we must have that $\phi(n) \in C_\delta^{n_0} \cup D_{\delta}^{n_0} \cup E^{n_0}$.

We now upper bound the number of instants $n$ which are associated to the same $\phi(n)$. Let $n,n^\prime \in B_{\epsilon}^{n_0}$ and $s = l_k(n_0,n)$. We see that $\phi(n^\prime) = \phi(n)$ implies either $\floor{l_{k}(n_0,n^\prime)/(2(\gamma + 1))} = \floor{l_{k}(n_0,n)/(2(\gamma + 1))}$ or $\floor{(3/2)l_{k}(n_0,n^\prime)/(\gamma + 1)} = \floor{(3/2)l_{k}(n_0,n)/(\gamma + 1)}$. Furthermore, $n^\prime \mapsto l_k(n_0,n^\prime)$ is incremented at time $n^\prime$. Hence for all $n \in B_{\epsilon}^{n_0}$:
	\eqs{
	| n^\prime \in B_{\epsilon}^{n_0} , \phi(n^\prime) = \phi(n)| \leq  2 \gamma (\gamma + 1).
	}
We have established that:
	\als{
	|B_{\epsilon}^{n_0}| &\leq l_0 + 2 \gamma (\gamma + 1) (|C_\delta^{n_0}| + |D_\delta^{n_0}|  + |E^{n_0}| ) \sk
	 &= 2(\gamma + 1) \log(1/\delta)/\delta^2 \sk
	 &+ 2 \gamma (\gamma + 1) (|C_\delta^{n_0}| + |D_\delta^{n_0}|  + |E^{n_0}| ).
	}
We complete the proof by providing bounds of the expected sizes of sets $C_\delta^{n_0}$, $D_\delta^{n_0}$ and $E^{n_0}$.

\medskip
\noindent
\underline{Bound of $\mathbb{E}[C_\delta^{n_0}]$}: Using Lemma~\ref{lem:deviation_ns} with $\Lambda(s) = \{ n \in C_\delta^{n_0}, l_k(n_0,n) =  s\}$, and by design of \ouralgosw: $t_k(n) \geq l_k(n_0,n)/(\gamma + 1) = s/(\gamma + 1)$. Since $|\Lambda(s)| \leq 1$ for all $s$, we have that:
	 \eqs{
	 \EE[ |C_\delta^{n_0}| ] \leq  O \Lp \frac{\log(\tau)}{\delta^2} \Rp.
	 }
\medskip
\noindent
\underline{Bound of $\mathbb{E}[D_{\delta}^{n_0}]$}: Using Lemma~\ref{lem:deviation_ns} with $\Lambda(s) = \{ n \in D_\delta^{n_0}, t_{k,k^\prime}(n_0,n) =  s\}$, and $|\Lambda(s)| \leq 1$ for all $s$, we have that:
	\eqs{
	 \EE[ |D_{\delta,k^\prime}^{n_0}| ] \leq  O \Lp \frac{\log(\tau)}{\delta^2} \Rp .
	 }

\medskip
\noindent
\underline{Bound of $\mathbb{E}[E^{n_0}]$}: By Lemma~\ref{lem:deviation_result} since $l_k(n) \leq \tau$:
\als{
	\PP[  n \in E^{n_0} ] &\leq 2 e \ceil{ \log(\tau) (\log(\tau) + c \log(\log(\tau)) ) } \sk 
	                      &\exp(- \log(\tau) + c \log(\log(\tau))) \sk
	 											& \leq \frac{4 e}{\tau \log(\tau)^{c-2}}.
}
Thus
	\eqs{
	\EE[ |E^{n_0}| ] \leq \frac{4 e}{(\log \tau)^{c-2}}.
	}

Putting the various bounds all together, we have: 
	\eqs{
	\EE[|{\cal L}_k(\Delta)[n_0,n_0+\tau] |] \leq O \Lp \frac{\log(\tau)}{\delta^2} \Rp,
	} 
for all $n_0 \in {\cal L}_k(\Delta)$, uniformly in $\delta$, which concludes the proof. \ep

\section{Proof of Proposition~\ref{th:unimodal_cont} }\label{sec:app_continuous}

The regret of UCB($\delta$) is defined as:
\eqs{R^{\pi}(T) \leq \sum_{k =1}^{\ceil{1/\delta}} \EE[t_k(T)] (\mu^* - \mu_k).}

 We separate the arms into three different sets. $\{ 1 ,\dots , \ceil{1/\delta} \} = A \cup B\cup C$, with: $A = \{k^* -1 , k^* , k^* + 1 \}$ the optimal arm and its neighbors, $B = \{ k: k  \notin A, (k-1) \delta \in [x^* - \delta_0,x^* + \delta_0] \}$ the arms which are not neighbors of the optimal arm, but are in $[x^* - \delta_0,x^* + \delta_0]$, and $C = \{ k:  (k-1) \delta \notin [x^* - \delta_0,x^* + \delta_0] \}$ the rest of the arms.

We consider $\delta < \delta_0/3$, so that $A \subset [x^* - \delta_0,x^* + \delta_0]$.
By our assumption on the reward function,  if $k \in A$, $| x^* - \delta (k-1)| \leq 2 \delta$ then  $| \mu^* - \mu_k | \leq C_2 (2\delta)^{\alpha}$. The regret is upper bounded by:
\eqs{ R^{\pi}(T) \leq T C_2 (2\delta)^{\alpha} + \sum_{k \in B \cup C} \EE[t_k(T)] (\mu^* - \mu_k).}
Using the fact that $\mu^* - \mu_{k^*} \leq C_2 \delta^\alpha$ and $\sum_{k =1}^{\ceil{1/\delta}} \EE[t_k(T)] \leq T$, the bound becomes:
\eqs{ R^{\pi}(T) \leq T C_2 (3\delta)^{\alpha} + \sum_{k \in B \cup C} \EE[t_k(T)] (\mu_{k^*} - \mu_k).}

By \cite{auer2002} (the analysis of UCB), for all $k$, $\EE[t_k(T)] \leq 8 \log(T) /(\mu_{k^*} - \mu_k)^2$. Replacing in the regret upper bound:
\eqs{ R^{\pi}(T) \leq T C_2 (3\delta)^{\alpha} +   \sum_{k \in B \cup C} 8 \log(T)/(\mu_{k^*} - \mu_k).}
If $k \in B$, $|\delta (k^*-1) - \delta (k-1) | \geq  \delta(|k^* - k| - 1)$, so $\mu_{k^*} - \mu_{k} \geq C_1 \delta^{\alpha}(|k^* - k| - 1)^\alpha$. If $k \in C$ , then $|\delta (k^*-1) - \delta (k-1) | \geq \delta_0/2$, so $\mu_{k^*} - \mu_{k} \geq C_1 (\delta_0/2)^\alpha$. So the regret for arms in $B \cup C$ reduces to:
\eqs{ R^{\pi}(T) \leq T C_2 (3\delta)^{\alpha} +    \frac{8 \log(T) \ceil{1/\delta} }{ C_1 (\delta_0/2)^\alpha}   + 2  \sum_{k = 1}^{\ceil{1/\delta}} \frac{8 \log(T)}{ C_1 (\delta k)^\alpha}.}
Using a sum-integral comparison: $ \sum_{k=1}^{\ceil{1/\delta}} k^{-\alpha} \leq \sum_{k=1}^{\ceil{1/\delta}} k^{-1} \leq  1 + \log( \ceil{1/\delta} )$, so that:
\als{ R^{\pi}(T) &\leq T C_2 (3\delta)^{\alpha} \sk &+  8 \log(T)   \left(   \frac{\ceil{1/\delta} }{ C_1 (\delta_0/2)^\alpha}  +    \frac{2 (1 + \log( \ceil{1/\delta} ))}{C_1 \delta^\alpha} \right).}

Setting $\delta =  ( \log(T)/\sqrt{T} )^{1/\alpha}$, the regret becomes:
\als{ R^{\pi}(T) \leq T C_2 (3^{\alpha})   ( \log(T)/\sqrt{T} ) +  \sk 8 \log(T)   \left( \frac{ \ceil{  (\sqrt{T}/\log(T))^{1/\alpha} }}{C_1 (\delta_0/2)^\alpha}  +    \frac{2 (1 + \log( T ))}{C_1  \log(T)/\sqrt{T} }    \right).}
we have used the fact that $\ceil{1/\delta} \leq T$.
\als{R^{\pi}(T)  \leq   C_2 (3^{\alpha}) \log(T) \sqrt{T} \sk +   8 \left(\frac{ \sqrt{T} + 1}{C_1 (\delta_0/2)^\alpha}  + \frac{2 \sqrt{T} (1 + \log( T ))}{C_1   } \right) 
}
Letting $T \to \infty$ gives the result:
\eqs{ \lim \sup_{T} R^{\pi}(T) / ( \sqrt{T} \log(T) ) \leq C_2 3^{\alpha} + 16 /C_1.}

\end{document}